\documentclass[10pt,twocolumn,letterpaper]{article}

\usepackage{cvpr}              

\usepackage{graphicx}
\usepackage{amsmath}
\usepackage{amssymb}
\usepackage{booktabs}
\usepackage{float}

\makeatletter
\@namedef{ver@everyshi.sty}{}
\makeatother
\usepackage{tikz}

\usetikzlibrary{arrows,shapes,calc,matrix,fit,backgrounds}
\usepackage{pgfplots}
\usepackage{pgfplotstable}
\pgfplotsset{compat=1.9}

\usepackage{xstring}

\usepgfplotslibrary{external}

\newcommand{\extdata}[1]{\input{#1}}
\IfBeginWith*{\jobname}{fig/extern/}{\finalcopy}{}

\tikzset{every mark/.append style={solid}}
\pgfplotsset{
	grid=both, width=\columnwidth, try min ticks=5,
	every axis/.append style={font=\scriptsize},
	every axis plot/.append style={thick,mark=none,mark size=1.2,tension=0.18},
	legend cell align=left, legend style={fill opacity=0.8},
}

\pgfplotsset{
	dash/.style={mark=o,dashed,opacity=0.7},
	dott/.style={mark=o,dotted,opacity=0.7},
}

\usepackage[pagebackref,breaklinks,colorlinks]{hyperref}

\usepackage[capitalize]{cleveref}
\crefname{section}{Sec.}{Secs.}
\Crefname{section}{Section}{Sections}
\Crefname{table}{Table}{Tables}
\crefname{table}{Tab.}{Tabs.}

\def\cvprPaperID{11263} 
\def\confName{CVPR}
\def\confYear{2022}

\newcommand{\citep}{\cite}
\newcommand{\citet}{\cite}

\begin{document}

\newcommand{\nn}[1]{\ensuremath{\text{NN}_{#1}}\xspace}

\newcommand{\citemiss}{\alert{[??]}\xspace}

\newcommand{\supe}[1]{^{\mkern-2mu(#1)}}
\newcommand{\dime}[1]{(#1)}

\def\l1{\ensuremath{\ell_1}\xspace}
\def\l2{\ensuremath{\ell_2}\xspace}

\newcommand*\OK{\ding{51}}

\newenvironment{narrow}[1][1pt]
	{\setlength{\tabcolsep}{#1}}
	{\setlength{\tabcolsep}{6pt}}

\newcommand{\comment} [1]{{\color{orange} \Comment     #1}} 
\newcommand{\commentout}[1]{}
\newcommand{\prm}[1]{_{#1}}

\newcommand{\alert}[1]{{\color{red}{#1}}}
\newcommand{\head}[1]{{\noindent\bf #1}}  
\newcommand{\equ}[1]{(\ref{equ:#1})\xspace}

\newcommand{\red}[1]{{\color{red}{#1}}}
\newcommand{\blue}[1]{{\color{blue}{#1}}}
\newcommand{\green}[1]{{\color{green}{#1}}}
\newcommand{\gray}[1]{{\color{gray}{#1}}}


\newcommand{\tran}{^\top}
\newcommand{\mtran}{^{-\top}}
\newcommand{\zcol}{\mathbf{0}}
\newcommand{\zrow}{\zcol\tran}

\newcommand{\ind}{\mathbbm{1}}
\newcommand{\expect}{\mathbb{E}}
\newcommand{\nat}{\mathbb{N}}
\newcommand{\zahl}{\mathbb{Z}}
\newcommand{\real}{\mathbb{R}}
\newcommand{\proj}{\mathbb{P}}
\newcommand{\prob}{\mathbf{Pr}}

\newcommand{\mif}{\textrm{if }}
\newcommand{\other}{\textrm{otherwise}}
\newcommand{\minimize}{\textrm{minimize }}
\newcommand{\maximize}{\textrm{maximize }}
\newcommand{\st}{\textrm{subject to }}

\newcommand{\id}{\operatorname{id}}
\newcommand{\const}{\operatorname{const}}
\newcommand{\sgn}{\operatorname{sgn}}
\newcommand{\var}{\operatorname{Var}}
\newcommand{\mean}{\operatorname{mean}}
\newcommand{\trace}{\operatorname{tr}}
\newcommand{\diag}{\operatorname{diag}}
\newcommand{\vect}{\operatorname{vec}}
\newcommand{\cov}{\operatorname{cov}}

\newcommand{\softmax}{\operatorname{softmax}}
\newcommand{\clip}{\operatorname{clip}}

\newcommand{\defn}{\mathrel{:=}}
\newcommand{\peq}{\mathrel{+\!=}}
\newcommand{\meq}{\mathrel{-\!=}}

\newcommand{\floor}[1]{\left\lfloor{#1}\right\rfloor}
\newcommand{\ceil}[1]{\left\lceil{#1}\right\rceil}
\newcommand{\inner}[1]{\left\langle{#1}\right\rangle}
\newcommand{\norm}[1]{\left\|{#1}\right\|}
\newcommand{\frob}[1]{\norm{#1}_F}
\newcommand{\card}[1]{\left|{#1}\right|\xspace}
\newcommand{\diff}{\mathrm{d}}
\newcommand{\der}[3][]{\frac{d^{#1}#2}{d#3^{#1}}}
\newcommand{\pder}[3][]{\frac{\partial^{#1}{#2}}{\partial{#3^{#1}}}}
\newcommand{\ipder}[3][]{\partial^{#1}{#2}/\partial{#3^{#1}}}
\newcommand{\dder}[3]{\frac{\partial^2{#1}}{\partial{#2}\partial{#3}}}

\newcommand{\wb}[1]{\overline{#1}}
\newcommand{\wt}[1]{\widetilde{#1}}

\def\xxssp{\hspace{-3pt}}
\def\xssp{\hspace{1pt}}
\def\ssp{\hspace{3pt}}
\def\msp{\hspace{5pt}}
\def\lsp{\hspace{12pt}}

\newcommand{\cA}{\mathcal{A}}
\newcommand{\cB}{\mathcal{B}}
\newcommand{\cC}{\mathcal{C}}
\newcommand{\cD}{\mathcal{D}}
\newcommand{\cE}{\mathcal{E}}
\newcommand{\cF}{\mathcal{F}}
\newcommand{\cG}{\mathcal{G}}
\newcommand{\cH}{\mathcal{H}}
\newcommand{\cI}{\mathcal{I}}
\newcommand{\cJ}{\mathcal{J}}
\newcommand{\cK}{\mathcal{K}}
\newcommand{\cL}{\mathcal{L}}
\newcommand{\cM}{\mathcal{M}}
\newcommand{\cN}{\mathcal{N}}
\newcommand{\cO}{\mathcal{O}}
\newcommand{\cP}{\mathcal{P}}
\newcommand{\cQ}{\mathcal{Q}}
\newcommand{\cR}{\mathcal{R}}
\newcommand{\cS}{\mathcal{S}}
\newcommand{\cT}{\mathcal{T}}
\newcommand{\cU}{\mathcal{U}}
\newcommand{\cV}{\mathcal{V}}
\newcommand{\cW}{\mathcal{W}}
\newcommand{\cX}{\mathcal{X}}
\newcommand{\cY}{\mathcal{Y}}
\newcommand{\cZ}{\mathcal{Z}}

\newcommand{\vA}{\mathbf{A}}
\newcommand{\vB}{\mathbf{B}}
\newcommand{\vC}{\mathbf{C}}
\newcommand{\vD}{\mathbf{D}}
\newcommand{\vE}{\mathbf{E}}
\newcommand{\vF}{\mathbf{F}}
\newcommand{\vG}{\mathbf{G}}
\newcommand{\vH}{\mathbf{H}}
\newcommand{\vI}{\mathbf{I}}
\newcommand{\vJ}{\mathbf{J}}
\newcommand{\vK}{\mathbf{K}}
\newcommand{\vL}{\mathbf{L}}
\newcommand{\vM}{\mathbf{M}}
\newcommand{\vN}{\mathbf{N}}
\newcommand{\vO}{\mathbf{O}}
\newcommand{\vP}{\mathbf{P}}
\newcommand{\vQ}{\mathbf{Q}}
\newcommand{\vR}{\mathbf{R}}
\newcommand{\vS}{\mathbf{S}}
\newcommand{\vT}{\mathbf{T}}
\newcommand{\vU}{\mathbf{U}}
\newcommand{\vV}{\mathbf{V}}
\newcommand{\vW}{\mathbf{W}}
\newcommand{\vX}{\mathbf{X}}
\newcommand{\vY}{\mathbf{Y}}
\newcommand{\vZ}{\mathbf{Z}}

\newcommand{\va}{\mathbf{a}}
\newcommand{\vb}{\mathbf{b}}
\newcommand{\vc}{\mathbf{c}}
\newcommand{\vd}{\mathbf{d}}
\newcommand{\ve}{\mathbf{e}}
\newcommand{\vf}{\mathbf{f}}
\newcommand{\vg}{\mathbf{g}}
\newcommand{\vh}{\mathbf{h}}
\newcommand{\vi}{\mathbf{i}}
\newcommand{\vj}{\mathbf{j}}
\newcommand{\vk}{\mathbf{k}}
\newcommand{\vl}{\mathbf{l}}
\newcommand{\vm}{\mathbf{m}}
\newcommand{\vn}{\mathbf{n}}
\newcommand{\vo}{\mathbf{o}}
\newcommand{\vp}{\mathbf{p}}
\newcommand{\vq}{\mathbf{q}}
\newcommand{\vr}{\mathbf{r}}
\newcommand{\vt}{\mathbf{t}}
\newcommand{\vu}{\mathbf{u}}
\newcommand{\vv}{\mathbf{v}}
\newcommand{\vw}{\mathbf{w}}
\newcommand{\vx}{\mathbf{x}}
\newcommand{\vy}{\mathbf{y}}
\newcommand{\vz}{\mathbf{z}}

\newcommand{\vone}{\mathbf{1}}
\newcommand{\vzero}{\mathbf{0}}

\newcommand{\valpha}{{\boldsymbol{\alpha}}}
\newcommand{\vbeta}{{\boldsymbol{\beta}}}
\newcommand{\vgamma}{{\boldsymbol{\gamma}}}
\newcommand{\vdelta}{{\boldsymbol{\delta}}}
\newcommand{\vepsilon}{{\boldsymbol{\epsilon}}}
\newcommand{\vzeta}{{\boldsymbol{\zeta}}}
\newcommand{\veta}{{\boldsymbol{\eta}}}
\newcommand{\vtheta}{{\boldsymbol{\theta}}}
\newcommand{\viota}{{\boldsymbol{\iota}}}
\newcommand{\vkappa}{{\boldsymbol{\kappa}}}
\newcommand{\vlambda}{{\boldsymbol{\lambda}}}
\newcommand{\vmu}{{\boldsymbol{\mu}}}
\newcommand{\vnu}{{\boldsymbol{\nu}}}
\newcommand{\vxi}{{\boldsymbol{\xi}}}
\newcommand{\vomikron}{{\boldsymbol{\omikron}}}
\newcommand{\vpi}{{\boldsymbol{\pi}}}
\newcommand{\vrho}{{\boldsymbol{\rho}}}
\newcommand{\vsigma}{{\boldsymbol{\sigma}}}
\newcommand{\vtau}{{\boldsymbol{\tau}}}
\newcommand{\vupsilon}{{\boldsymbol{\upsilon}}}
\newcommand{\vphi}{{\boldsymbol{\phi}}}
\newcommand{\vchi}{{\boldsymbol{\chi}}}
\newcommand{\vpsi}{{\boldsymbol{\psi}}}
\newcommand{\vomega}{{\boldsymbol{\omega}}}

\newcommand{\rLambda}{\mathrm{\Lambda}}
\newcommand{\rSigma}{\mathrm{\Sigma}}

\def\onedot{.\xspace}
\def\eg{\emph{e.g}\onedot} \def\Eg{\emph{E.g}\onedot}
\def\ie{\emph{i.e}\onedot} \def\Ie{\emph{I.e}\onedot}
\def\cf{\emph{cf}\onedot} \def\Cf{\emph{C.f}\onedot}
\def\etc{\emph{etc}\onedot}
\def\vs{\emph{vs}\onedot}
\def\wrt{w.r.t\onedot} \def\dof{d.o.f\onedot}
\def\etal{\emph{et al}\onedot}

\newcommand{\std}[1]{{\tiny\textpm{}{#1}}}

\makeatother

\newcommand{\exstd}[1]{{\tiny\textit{}{$\pm$#1}}}   

\title{Learning with Neighbor Consistency for Noisy Labels}

\author{
Ahmet Iscen$^1$ \ \ \ \ Jack Valmadre\thanks{Work done at Google.}\ \ $^2$\ \ \ \ Anurag Arnab$^1$\ \ \ \ Cordelia Schmid$^{1}$\\
{\fontsize{11}{13}\selectfont$^1$Google Research\ \ \ \ \ \ $^2$AIML, University of Adelaide}\\
}

\maketitle

\begin{abstract}

Recent advances in deep learning have relied on large, labelled datasets to train high-capacity models.
However, collecting large datasets in a time- and cost-efficient manner often results in label noise.
We present a method for learning from noisy labels that leverages similarities between training examples in feature space, encouraging the prediction of each example to be similar to its nearest neighbours.
Compared to training algorithms that use multiple models or distinct stages, our approach takes the form of a simple, additional regularization term.
It can be interpreted as an inductive version of the classical, transductive label propagation algorithm.  
We thoroughly evaluate our method on datasets evaluating both synthetic (CIFAR-10, CIFAR-100) and realistic (mini-WebVision, WebVision, Clothing1M, mini-ImageNet-Red) noise, and achieve competitive or state-of-the-art accuracies across all of them.
\footnote{The code is available at \url{https://github.com/google-research/scenic/tree/main/scenic/projects/ncr}}
\end{abstract}

\section{Introduction}

While deep learning can achieve unprecedented accuracy in image classification tasks, it requires a large, supervised dataset that is often expensive to obtain.
Unsupervised and semi-supervised learning seek to alleviate this requirement by incorporating unlabelled examples.
However, these approaches cannot take advantage of the various sources of \emph{noisy} labels in the modern world, such as images with hashtags in social media or images contained in webpages retrieved by a textual query.
Training algorithms that are robust to label noise are therefore highly attractive for deep learning.

The dominant approach to learning with noisy labels in recent work is to use the predictions of the model itself to reject or modify training examples (\eg~\cite{reed2014training, tanaka2018joint, arazo2019unsupervised, yang2020webly, liu2020early, nguyen2020self, han2018coteaching, li2020dividemix}).
This is inherently risky due to the ability of deep networks to fit arbitrary labels~\cite{zhang2016understanding} and it is important to take significant measures against overfitting.
Moreover, this paradigm often leads to complicated training procedures, such as maintaining multiple models or alternating between updating the model and updating the training set.

\begin{figure*}[t]
\begin{center}
\includegraphics[width=0.7\textwidth]{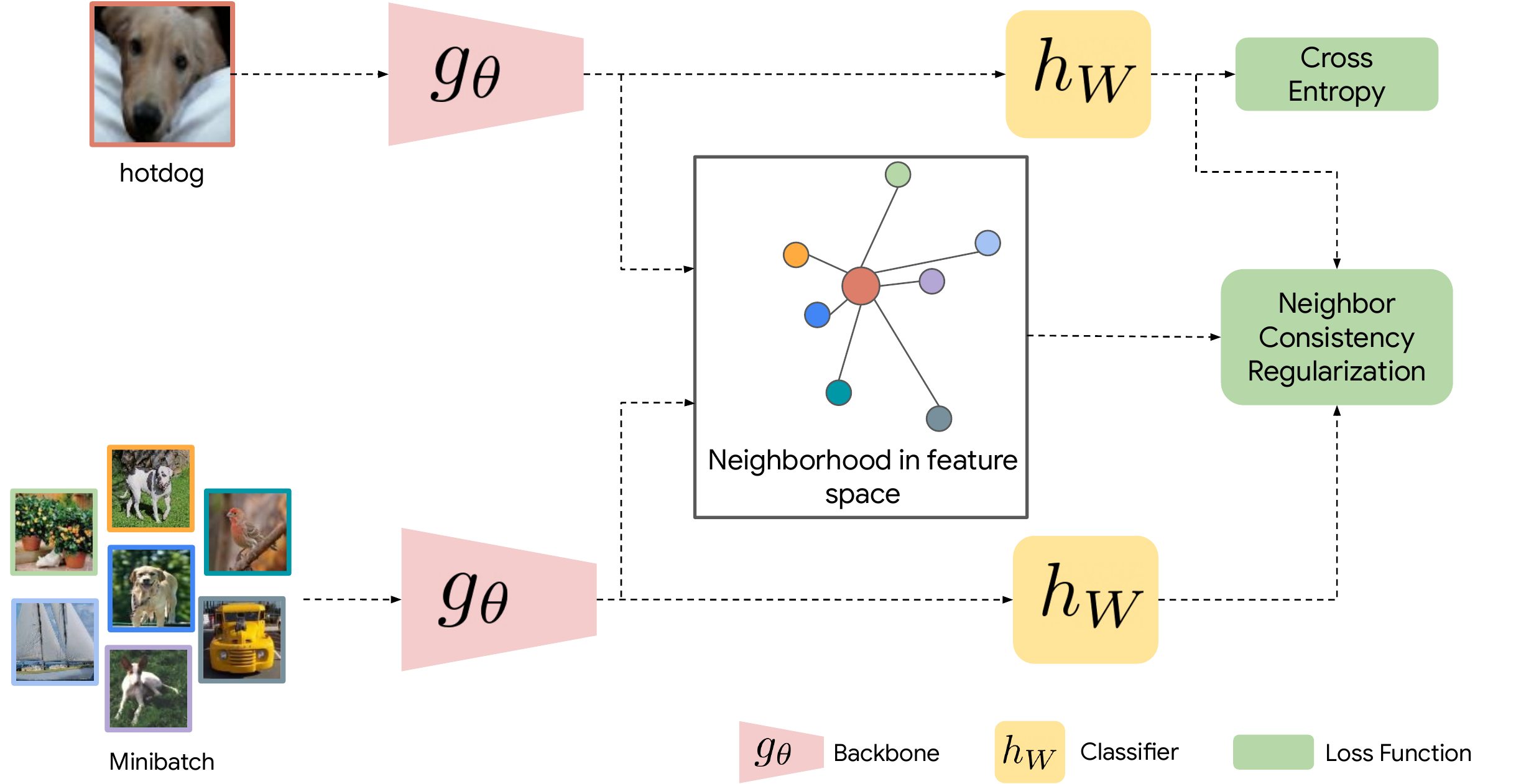}
\end{center}
\caption{
To address the problem of noisy labels in the training set, we propose Neighbor Consisteny Regularization. This regularizer encourages examples with similar feature representations to have similar outputs, thus mitigating the impact of training examples with incorrect labels.
\label{fig:overview}
}
\vspace{-0.5\baselineskip}
\end{figure*}

This paper proposes Neighbor Consistency Regularization (NCR) for the specific problem of learning with noisy labels, illustrated in Figure~\ref{fig:overview}.
Rather than adopting model predictions as pseudo-labels, NCR introduces an additional consistency loss that encourages each example to have similar predictions to its neighbors.
Concretely, the neighbor consistency loss penalizes the divergence of each example's prediction from a weighted combination of its neighbors' predictions, with the weights determined by their similarity in feature space.
The motivation of NCR is to enable incorrect labels to be improved or at least attenuated by the labels of their neighbors, relying on the assumption that the noise is sufficiently weak or unstructured so as not to overwhelm the correct labels.
Compared to the popular approach of bootstrapping the model predictions~\cite{reed2014training}, NCR can be seen as instead bootstrapping the learned feature representation, which may reduce its susceptibility to overfitting and improve its stability at random initialization. 

NCR is inspired by label propagation algorithms for semi-supervised learning~\cite{ZBL+03, KW17, iscen2019label}, which seek to transfer labels from supervised examples to neighboring unsupervised examples according to their similarity in feature space.
However, whereas label propagation is typically performed in a batch setting over the entire dataset, our method effectively performs label propagation online within mini-batches during stochastic gradient descent.
This results in a simple, single-stage training procedure.
Moreover, whereas existing methods for label propagation represent \emph{transductive} learning in that they only produce labels for the specific examples which are provided during training, NCR can be understood as an \emph{inductive} form of label propagation in that it produces a model which can later be applied to classify unseen examples.

The key contributions of the paper are as follows.
\begin{itemize}
\item We propose Neighbor Consistency Regularization, a novel loss term for deep learning with noisy labels that encourages examples with similar feature representations to have similar predictions.
\item We verify empirically that NCR achieves better accuracy than several important baselines at a wide range of noise levels from both synthetic and real distributions, and is complementary to the popular regularization technique of mixup \cite{zhang2017mixup}.
\item We demonstrate that NCR achieves competitive or state-of-the-art accuracies on datasets evaluating both synthetic (CIFAR-10 and CIFAR-100) and realistic (mini-WebVision \cite{li2020dividemix}, mini-ImageNet-Red \cite{jiang2020beyond}, Clothing1M \citep{xiao2015learning}) noise scenarios.
\end{itemize}

\section{Related work}
\label{sec:related_work}

This section reviews related work with a focus on image classification using deep learning and draws comparisons to our proposed approach.
Note that, while there is a significant body of work examining the problem of noisy labels under the assumption that a small, trusted set of clean examples is available, this paper considers the variant where this is not the case.

\head{Regularization.}
Regularization has an integral role under label noise as it limits the ability of a high-capacity model to fit arbitrary labels.
Indeed, regularization alone can provide considerable robustness to label noise and many algorithms couple an effective regularization strategy with an explicit noise-handling technique to further robustness.
One particularly effective form of regularization is mixup augmentation~\citet{zhang2017mixup}, which generates additional examples by linearly interpolating between pairs of examples in both the image and label space.
It was originally demonstrated to provide robustness to synthetic label corruption on CIFAR-10. 

Beyond the choice of regularization, there are numerous possible approaches to handle noisy labels.
Early investigation into deep learning with label noise concentrated on either estimation of the label transition matrix~\citep{sukhbaatar2014training,PRK+17} or loss functions that are robust to outliers~\citep{reed2014training,ghosh2017robust,xu2019ldmi}.
More recently, numerous methods have been proposed that re-weight or re-label the examples that are believed to have incorrect labels.
We briefly review this body of work in the following paragraphs.

\head{Model predictions for noisy labels.}
Many methods have used the predictions of the model itself during training to either generate pseudo-labels or to identify incorrect examples (or both).
For deep learning, this often leverages the phenomenon that correct labels tend to be approximated earlier than incorrect ones \citep{liu2020early}.
Reed~\etal~\citet{reed2014training} first proposed the bootstrapping loss, in which a fixed linear combination of the annotated label and the current prediction are adopted as the regression target for each example.
Our method can be considered a bootstrapping approach which uses the learned \emph{similarity} rather than the actual predictions.
Extensions of~\citet{reed2014training} include setting the proportion of annotated and predicted labels adaptively using the model confidence~\citep{arazo2019unsupervised, yang2020webly} and averaging model weights and predictions over time~\citep{liu2020early, nguyen2020self}.
Lukasik~\etal~\citet{lukasik2020does} showed that some robustness to noise could be obtained by label smoothing, which instead interpolates between the annotated label and a uniform distribution.
To avoid overfitting, several works have proposed to split the training set in half and train two models, with each used to assess the training examples of the other~\citep{han2018coteaching, li2020dividemix}.  

\head{Identifying noisy examples using neighbors.}
Whereas our method, in effect, uses neighbors to modify the supervision of each example, several approaches have rather used neighbors with the aim of identifying and eliminating noisy examples. 
Both \citet{wang2018iterative} and \citet{guo2018curriculumnet} proposed to identify and re-weight examples with incorrect labels using the local density of examples with the same annotation.
Wu~\etal~\citet{wu2020topological} constructed a $k$-NN graph and kept only the examples which constituted the core of the largest connected component per class.
Bahri~\etal~\citet{bahri2020deep} eliminated examples whose labels did not agree with the prediction of a $k$-NN classifier.  
Multi-Objective Interpolation Training (MOIT)~\cite{ortego2020multi} identifies mislabelled examples by comparing the predictions of each example to the average prediction of its neighbors and then replaces the least reliable labels with the model's current predictions.
MOIT+ incorporates an additional stage of semi-supervised learning, in which the labels assumed to be noisy, are discarded.

\head{Label propagation.}
Variants of label propagation~\citep{ZBL+03} have been applied to various computer vision tasks, including retrieval~\citep{DB13,ITA+17}, semi-supervised learning~\citep{KW17, iscen2019label} and few-shot learning~\citep{DSH+18, liu2018learning, rodriguez2020embedding}.
For semi-supervised learning in particular, \citet{iscen2019label} used it to obtain labels for unsupervised examples based on their neighbors in feature space.
While most methods perform label propagation over large graphs containing many examples in a batch fashion, \citet{liu2018learning} and \citet{rodriguez2020embedding} considered label propagation within stochastic gradient descent for episodic few-shot learning~\citep{VBL+16}.
This is a meta-learning approach which seeks a suitable feature representation with which to \emph{later} perform label propagation during meta-testing.
It is not directly applicable for learning with noisy labels and it does not scale to large sets of examples.

For learning with noisy labels, although neighbors have often been used to \emph{identify} mislabelled examples (see above), few works have considered the use of neighbors to generate pseudo-labels.
Iscen~\etal~\citet{iscen2020graph} used graph convolution to propagate labels from a small set of examples with trusted labels to a large set of examples with noisy labels.
By comparison, our approach does not require a clean set and is inductive in nature.
Perhaps the most similar method to our own is that of \citet{yang2020webly}, in which graph filtering is applied once to the global graph of all examples to refine the predictions of an initial model.
In contrast, our consistency loss encourages similar examples to have similar labels throughout training and is much simpler to implement.

\head{Consistency.}
The idea of constraining a network to produce consistent outputs as a way to leverage unlabelled data has appeared in several previous works.
ICT~\citet{verma2019interpolation} and Mixmatch~\citet{berthelot2019mixmatch} proposed variants of mixup for semi-supervised learning in which predictions replace the labels for unsupervised examples. 
Xie~\etal~\citet{xie2020unsupervised} introduced Unsupervised Data Augmentation for semi-supervised image classification, where a model is encouraged to be robust to label-preserving transformations even when labels are not available by minimizing the divergence between predictions for transformed and non-transformed images.
Most relevant to our work, \citet{englesson2021generalized} used prediction consistency with respect to image transformations for the express purpose of learning with noisy labels.
While these forms of consistency are effective regularizers, neighbor consistency offers the ability to transfer supervision directly to mislabelled examples.
\section{Preliminaries}

We first define our notation and formulate the task of learning with noisy labels.
We then describe label propagation, a graph-based semi-supervised learning method designed to work with closed datasets.

\head{Problem formulation.}
We assume a dataset defined by $X \defn \{x_1, \ldots, x_n\}$.
Each example, \eg an image, $x_i$, has a corresponding true label $\tilde{y}_i \in \mathcal{C}$.
In our task, some of the labels, $y_i$, are noisy: $y_i \neq \tilde{y}_i$ and do not correctly reflect the visual content of the example $x_i$.
During training, we do not know whether $y_i$ is noisy ($y_i \neq \tilde{y}_i$) or clean ($y_i = \tilde{y}_i$).
Our goal is to learn a model with the highest accuracy on the true labels, $\tilde{y}$, although an unknown number of labels in our training set are noisy.

We learn a convolutional neural network for classification. 
The network, denoted by  $f\prm{\theta,W}: \cX \rightarrow \real^c$, takes a dataset example $x_i$ as the input and outputs \emph{logits} for softmax classification.
Its two learnable variables $\theta$ and $W$ correspond to the feature extractor and classifier, respectively. 
The feature extractor maps an image $x_i$ to a $d$-dimensional vector $\vv_i \defn g\prm{\theta}(x_i) \in \real^d$.
The classifier maps the $d$-dimensional vector to class scores $\vz_i \defn h\prm{W}(\vv_i) \in \real^c$.
Typically, the network parameters are learned by minimizing a loss function for supervised classification:
\begin{equation}
L_{\text{S}}( X, Y; \theta, W) \defn \frac{1}{m}\sum_{i=1}^m \ell \left( \sigma(\vz_i), y_i\right),
\label{eq:loss}
\end{equation}
where $X$ and $Y$ correspond to the set of examples in the mini-batch, $m = |X| = |Y|$ denotes the size of the mini-batch, $\sigma$ is the softmax function and $\ell(\mathbf{q}, \mathbf{p})$ is the cross-entropy loss function for predictions~$\mathbf{q}$.
When the target distribution~$\mathbf{p}$ is a single label~$y \in \mathcal{C}$, we adopt the short-hand $\ell(\mathbf{q}, y) = \ell(\mathbf{q}, \boldsymbol{\delta}_{y})$ for cross-entropy with a one-hot vector~$\boldsymbol{\delta}_{y}$.

\head{Label propagation} is a graph-based technique used in semi-supervised learning~\citep{ZBL+03}.
We assume that we are given labeled and unlabeled examples in a dataset, and that the dataset is also defined by a graph which is either given or created from the k-NN of each example~\citep{DB13}.
This method spreads the label information of each node to the other nodes based on the connectivity in the graph.
This process is repeated until a global equilibrium state is achieved.
Finally, unlabeled examples are assigned to the class from which they have received the most information.
Formally, for a graph of dataset $X$, represented by an affinity matrix $W$, where $W_{ij} = \text{similarity}(x_i, x_j)$, Zhou~\etal~\cite{ZBL+03} show that label propagation can be computed by minimizing
\begin{align}
\mathcal{Q}(P) &=  \frac{1}{2} \mu \sum_{i=1}^n \Vert P_i - Y_i \Vert^2 \nonumber\\
&+ \frac{1}{2} \sum_{i,j=1}^{n} W_{ij} \Big\Vert \frac{1}{\sqrt{D_{ii}}}P_i - \frac{1}{\sqrt{D_{jj}}}P_j \Big\Vert^2 \enspace ,
\label{eq:lpreg}
\end{align}
where $D$ is the degree matrix (a diagonal matrix with entries $D_{i i} = \sum_{j} W_{i j}$),
$P \in \mathbb{R}^{n\times c}$ is the matrix of classification predictions,
$Y \in \mathbb{R}^{n\times c}$ is the matrix of one-hot labels for all examples and
$\mu$ is a regularization parameter.
This objective function comprises two terms: a \emph{fitting constraint}, which encourages the classification of each point to their assigned label, and a \emph{smoothing} term, which encourages the outputs of nearby points in the graph to be similar.

One of the main limitations of label propagation is its \emph{transductive} property.
In transductive learning, the goal is to classify \emph{seen} unlabeled examples.
This is different to \emph{inductive} learning, which learns a generic classifier to classify any unseen data.
To apply label propagation on new test examples, a new graph $W$ needs to be constructed each time a test example is seen.
This makes it inefficient in practice.

Another requirement for label propagation is that the feature space needs to be fixed to compute the affinity matrix $W$.
This requires the feature extractor to be learned beforehand, potentially from the noisy data.
Existing work~\citep{iscen2019label} has tried to overcome this issue by alternating between optimizing the feature space, and performing label propagation.
However, this does not directly enforce smoothness, as the optimization of two components are done separately.

Our goal is to overcome the limitations of label propagation by 1) adapting it to an inductive setting 2) applying the smoothness constraint directly during optimization.
In Section~\ref{sec:method}, we propose a simple and efficient approach which generalizes label propagation by enforcing smoothness in the form of a regularizer. 
As a result, we avoid constructing an explicit graph to propagate the information, and inference can be performed on any unseen test example.  

\section{Method} 
\label{sec:method}

We now present our method \emph{Neighbor Consistency Regularization} and compare it to classical label propagation.
We then highlight its relationship to similar, online techniques.

\subsection{Neighbor Consistency Regularization}

When learning with noisy labels, the network is prone to overfit, or \emph{memorize}, the mapping from $x_i$ to a noisy label $y_i$ for the training data \citep{liu2020early}.
This behavior typically results in a non-optimal classification performance in a clean evaluation set, as the network does not generalize well.

To overcome this issue, we propose \emph{Neighbor Consistency Regularization} (NCR). 
Our main assumption is that the over-fitting occurs less dramatically before the classifier $h\prm{W}$.
This is supported by MOIT~\cite{ortego2020multi}, which shows that feature representations are robust enough to discriminate between noisy and clean examples when training a network.
With that assumption, we can design a smoothness constraint similar to label propagation~\eqref{eq:lpreg} when training the network.
The overview of our method is shown in Figure~\ref{fig:overview}.

Let us define the \emph{similarity} between two examples by the cosine similarity of their feature representations, \ie $s_{i,j} = \text{cos}(\vv_i,\vv_j) = \vv_i^{T} \vv_j / (\|\vv_{i}\| \|\vv_{j}\|)$.
Note that the feature representations contain non-negative values when obtained after a ReLU non-linearity, and therefore the cosine similarity is bounded in the interval $[0, 1]$. 
Our goal is to enforce \emph{neighbor consistency regularization} by leveraging the structure of the feature space produced by $g\prm{\theta}$ to enhance the classifier $h\prm{W}$.
More specifically, $h\prm{W}(\vv_i)$ and $h\prm{W}(\vv_j)$ should behave similarly if $s_{i,j}$ is high, regardless of their labels $y_i$ and $y_j$.
This would prevent the network from over-fitting to an incorrect mapping between an example $x_i$ and a label $y_i$, if either (or both) $y_i$ and $y_j$ are noisy.

To enforce NCR, we design an objective function which minimizes the distance between logits $\vz_i$ and $\vz_j$, if the corresponding feature representations $\vv_i$ and $\vv_j$ are similar:
\begin{align}
 &L_{\text{NCR}}( X, Y; \theta, W) \defn \nonumber\\ &\frac{1}{m}\sum_{i=1}^m  D_\text{KL} \bigg( \sigma(\vz_i/T) \;\bigg\Vert\; 
 \sum_{j \in  \nn{k}(\vv_i) } \frac{s_{i, j}}{\sum_{k} s_{i, k}} \cdot \sigma(\vz_j/T) \bigg),
\label{eq:lcloss}
\end{align}

where $D_\text{KL}$ is the KL-divergence loss to measure the difference between two distributions, $T$ is the temperature and $\nn{k}(\vv_i)$ denotes the set of $k$ nearest neighbors of $i$ in the feature space.
We set $T=2$ throughout our experiments.
We normalize the similarity values so that the second term of the KL-divergence loss remains a probability distribution. 
We set the self-similarity $s_{i,i} =0$ so that it does not dominate the normalized similarity.
Gradients will be back-propagated to all inputs.

The objective~\eqref{eq:lcloss} ensures that the output of $x_i$ will be consistent with the output of its neighbors regardless of its potentially noisy label $y_i$.
We combine it with the supervised classification loss function~\eqref{eq:loss} to obtain the final objective to minimized during the training:
\begin{align}
L( X, Y; \theta, W) &\defn  (1-\alpha) \cdot L_{\text{S}}( X, Y; \theta, W) \nonumber\\ &+ \alpha \cdot L_{\text{NCR}}( X, Y; \theta, W),
\label{eq:finalloss}
\end{align}
where the hyper-parameter $\alpha \in [0, 1]$ controls the impact of the each loss term.
Similar to label propagation, the final loss objective~\eqref{eq:finalloss} has two terms. 
The first term is the classification loss term $L_{\text{S}}$.
This is analogous to the fitting constraint in~\eqref{eq:lpreg}.
The second term is the NCR loss $L_{\text{NCR}}$, which is also similar to the \emph{smoothness} constraint in~\eqref{eq:lpreg}.
We find that it sometimes helps to train the network with $\alpha = 0$ for several epochs (denoted by $e$ in our experiments) before enabling the NCR term.

However, the main difference between label propagation and our method is that label propagation applies smoothness based on the graph edges $W_{ij}$ computed over the entire dataset.
On the other hand, our method is online, and does not require a global graph $W$. 
We enforce the NCR through the local neighborhood as the feature space is being learned.
As a result, our method does not require a learned feature representation with noisy examples.
It enriches the learned feature representation by reducing the negative impact of noisy examples.

Compared to standard training, NCR incurs an additional computational cost of order $O(m^2 (d + c))$ where $m$ is the batch size, $d$ is the feature dimension and $c$ is the number of classes.
This arises in the computation of the similarity values and weighted predictions in~\eqref{eq:lcloss}. 
However, this operation is relatively fast to compute for moderate values of~$m$ because it is a dense matrix multiplication, for which modern GPUs are optimized.

\subsection{Relation to other methods}
\label{sec:relation}

\head{Bootstrapping} introduces an additional loss that adopts the model's own predictions as labels~\citep{reed2014training}.
Its motivation is to discourage the model from overfitting to examples which are difficult to fit.
The overall loss combines the supervised loss and the bootstrap loss in fixed proportion:
\begin{align}
L_{\text{B}}(X, Y; \theta, W) &\defn \frac{1}{m} \sum_{i = 1}^{m} (1 - \alpha) \cdot \ell(\sigma(\vz_{i}), y_{i}) \nonumber\\ &+ \alpha \cdot \ell(\sigma(\vz_{i}),\ \sigma_{\text{B}}(\vz_{i})),
\end{align}
where $\sigma_{\text{B}}$ is the bootstrap activation function, which may be argmax or softmax (with optional temperature).
NCR can be understood as bootstrapping from the neighborhood structure induced by the representation rather than from the model's predictions.
This eliminates the dependency on the classifier parameters~$W$, which may be particularly advantageous as it has been shown that linear models can fit random labels given a sufficiently high-dimensional representation~\citep{liu2020early}.

\head{Label smoothing} is a regularization method~\citep{szegedy2016rethinking} that mixes the ground truth labels with a uniform distribution.
It has been shown capable of denoising corrupted labels~\citep{lukasik2020does}.
Under label smoothing, the supervised classification loss function becomes:
\begin{align}
  L_{\text{LS}}( X, Y; \theta, W) &\defn \frac{1}{m}\sum_{i=1}^m (1-\alpha) \cdot \ell \left( \sigma(\vz_i), y_i\right) \nonumber\\ &+ \alpha \cdot \ell \big( \sigma(\vz_i),  \tfrac{1}{C} \mathbf{1} \big).
\label{eq:lsloss}
\end{align}
Note that the linear combination of losses is equivalent to a linear combination of labels due to the linearity of $\ell(\vq, \vp) = -\vp^{T} \log \vq$ with respect to~$\vp$.
Our method can be considered a modified version of label smoothing where the uniform distribution is replaced with a distribution defined by the neighboring examples.
It would collapse to label smoothing if the neighbors were random or if a high temperature~$T$ were used.

\head{Mixup} \citep{zhang2017mixup} bears some resemblance to our method in that it takes a convex combination of labels.
However, mixup uses this combination as the regression target for a novel example obtained as the convex combination of the inputs, whereas NCR uses it as the target for an existing example.

\section{Experiments}

\subsection{Experimental setup}
\label{sec:expsetup}

We first perform ablation studies on datasets with synthetic noise, where the noise level can be varied, before considering datasets with organic noise, where the noise level is fixed and unknown.
For experiments with synthetic noise, we use the standard variants of CIFAR-10 and -100 \citep{KH09} as well as mini-ImageNet with ``Blue'' and ``Red'' noise \citep{jiang2020beyond}.
The CIFAR and mini-ImageNet-Blue datasets are contaminated with uniform label noise, whereas mini-ImageNet-Red is constructed by replacing some examples in each class with false positives from an image search engine, representing more realistic noise.
For experiments with organic noise, we use mini-WebVision~\citep{li2020dividemix}, WebVision~\cite{li2017webvision} and Clothing1M \citep{xiao2015learning}.
All datasets include a clean validation and/or final evaluation set.
Exhaustive implementation details and hyperparameters are also included in the supplementary.

\begin{figure*}
\newenvironment{customlegend}[1][]{%
    \begingroup
    \csname pgfplots@init@cleared@structures\endcsname
    \pgfplotsset{#1}%
}{%
    \csname pgfplots@createlegend\endcsname
    \endgroup
}%

\def\addlegendimage{\csname pgfplots@addlegendimage\endcsname}

\centering
\input{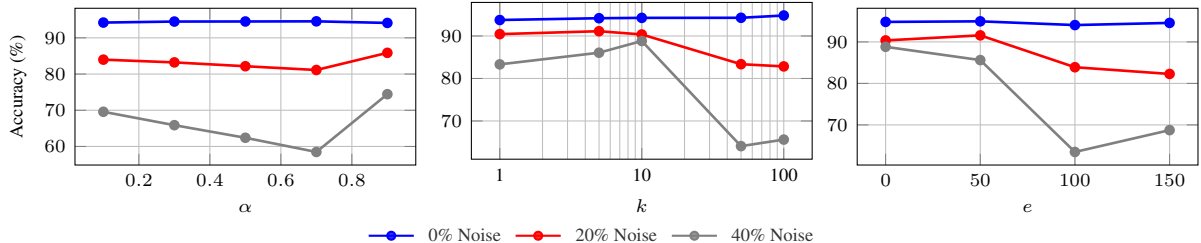}
\begin{tabular}{@{\xxssp}c@{\xssp}c@{\xxssp}c@{\xxssp}}
{

\begin{tikzpicture}
    \tikzstyle{every node}=[font=\scriptsize]
\begin{axis}[%
  width=0.35\textwidth,
  height=0.21\textwidth,
  xlabel={$\alpha$},
    grid=both,
  ylabel= {Accuracy (\%)},
  xminorticks=false,
  legend cell align={left},
  legend pos=outer north east,
  legend style={at={(0.5,1.5)},anchor=north, font =\scriptsize, fill opacity=0.8, row sep=-2.5pt, legend columns=-1},
]

  \addplot[color=blue,     solid, mark=*,  mark size=1.5, line width=1.0] table[x=alpha, y expr={\thisrow{zero}*100}] \cifaralpha;
  \addplot[color=red,     solid, mark=*,  mark size=1.5, line width=1.0] table[x=alpha, y expr={\thisrow{twenty}*100}] \cifaralpha;
  \addplot[color=gray,     solid, mark=*,  mark size=1.5, line width=1.0] table[x=alpha, y expr={\thisrow{forty}*100}] \cifaralpha;

\end{axis}
\end{tikzpicture}

}
&

{
\begin{tikzpicture}
    \tikzstyle{every node}=[font=\scriptsize]
\begin{axis}[%
  width=0.35\textwidth,
  height=0.21\textwidth,
  xlabel={$k$},
    xmode=log,
    grid=both,
  xminorticks=false,
  xtick={1, 10, 100},
  xticklabels={1, 10, 100},  
  legend cell align={left},
  legend pos=outer north east,
  legend style={at={(0.5,1.5)},anchor=north, font =\scriptsize, fill opacity=0.8, row sep=-2.5pt, legend columns=-1},
]

  \addplot[color=blue,     solid, mark=*,  mark size=1.5, line width=1.0] table[x=k, y expr={\thisrow{zero}*100}] \cifark;
  \addplot[color=red,     solid, mark=*,  mark size=1.5, line width=1.0] table[x=k, y expr={\thisrow{twenty}*100}] \cifark;
  \addplot[color=gray,     solid, mark=*,  mark size=1.5, line width=1.0] table[x=k, y expr={\thisrow{forty}*100}] \cifark;
\end{axis}
\end{tikzpicture}
}
&

{
\begin{tikzpicture}
    \tikzstyle{every node}=[font=\scriptsize]
\begin{axis}[%
  width=0.35\textwidth,
  height=0.21\textwidth,
  xlabel={$e$},
    grid=both,
  xminorticks=false,
  legend cell align={left},
  legend pos=outer north east,
  legend style={at={(0.5,1.5)},anchor=north, font =\scriptsize, fill opacity=0.8, row sep=-2.5pt, legend columns=-1},
]

  \addplot[color=blue,     solid, mark=*,  mark size=1.5, line width=1.0] table[x=e, y expr={\thisrow{zero}*100}] \cifarepochs;
  \addplot[color=red,     solid, mark=*,  mark size=1.5, line width=1.0] table[x=e, y expr={\thisrow{twenty}*100}] \cifarepochs;
  \addplot[color=gray,     solid, mark=*,  mark size=1.5, line width=1.0] table[x=e, y expr={\thisrow{forty}*100}] \cifarepochs;

\end{axis}
\end{tikzpicture}

}

\end{tabular}

\centering
\vspace{-0.5\baselineskip}
\begin{tikzpicture}
\scriptsize
\begin{customlegend}[legend columns=4,legend style={align=left,draw=none,column sep=1ex},
        legend entries={0\% Noise,
                        20\% Noise,
                        40\% Noise,
                        }]
        \addlegendimage{color=blue, solid, mark=*, mark size=1.5}
        \addlegendimage{color=red, solid, mark=*, mark size=1.5}
        \addlegendimage{color=gray, solid, mark=*, mark size=1.5}
        \end{customlegend}
\end{tikzpicture}
\vspace{-1\baselineskip}
\caption{\textbf{Ablation study.} Impact of hyperparameters $\alpha$, $k$ and $e$ are evaluated on the CIFAR-10 validation set with ResNet-18.
\label{fig:ablation} 
}
\vspace{-0.5\baselineskip}
\end{figure*}

\subsection{Ablation study}

We first study the effect of the key hyperparameters of NCR at different noise levels using the CIFAR-10 validation set.
Specifically, we investigate the impact of~$\alpha$, which controls the strength of the NCR term in~\eqref{eq:finalloss}, the number of neighbors $k$ and the number of initialization epochs $e$.  
We first set the number of neighbors $k = m$ (the batch size) and the number of initialization epochs to zero while varying~$\alpha$.
Subsequently, we select the optimal~$\alpha$ for each noise level and vary $k$.
Finally, we adopt the optimal $k$ and vary $e$.

Figure~\ref{fig:ablation} shows the validation accuracy for different noise ratios.
The performance remains relatively stable over different hyperparameters for $0\%$ and $20\%$ noise ratios.
It is optimal to set $\alpha$ to be high (0.9) for any non-zero noise ratio, indicating that the greater influence of NCR benefits these settings.
Similarly, smaller $k$ (\eg $k=10$) leads to higher accuracy when the noise ratio is above $0\%$.
We also observe that $e=0$ performs better for higher noise ratios (\ie $40\%$ and $80\%$).
This shows that the NCR needs to be enabled early in training to prevent the network from memorizing the noisy labels when the noise ratio is high.

\subsection{Baseline comparison}
\label{sec:baselinecomp}

\begin{table*}
  \caption{\textbf{Baseline and oracle comparison.} Classification accuracy is reported on the mini-ImageNet-\{Blue, Red\} datasets with the ResNet-18 architecture for each individual noise ratio ($0\%$, $20\%$, $40\%$, $80\%$).
  We present the mean accuracy and standard deviation from five trials.
  The oracle model is trained on only the known, clean examples in the training set using a cross-entropy loss.
  }
  \scriptsize
\begin{center}
\begin{tabular}{@{\msp}l@{\msp}@{\msp}c@{\msp}c@{\msp}c@{\msp}c@{\msp}@{\lsp}c@{\msp}c@{\msp}c@{\msp}c@{\msp}}
\toprule
				& \multicolumn{4}{c}{mini-ImageNet-Blue} & \multicolumn{4}{c}{mini-ImageNet-Red} \\
\midrule 
Method 											& $0\%$ 				& $20\%$ 				& $40\%$ 					& $80\%$ 					& $0\%$				& $20\%$ 				& $40\%$ 					& $80\%$				\\
\midrule
\multicolumn{9}{c}{\textbf{\textsc{Baselines}}}  \\ \midrule
Standard							            & 72.7\exstd{0.4}				& 63.4\exstd{0.4}			& 55.9\exstd{0.5} 			& 13.4\exstd{0.4}				& 70.9\exstd{0.5}		& 66.9\exstd{0.4}			& 63.0\exstd{0.3}				& 49.3\exstd{0.6} \\
Mixup 											& 72.6\exstd{0.2}				& 66.5\exstd{0.4}			& 59.4\exstd{0.8}			& 7.8\exstd{0.4}				& 70.5\exstd{0.2}		& 67.6\exstd{0.3}			& 63.8\exstd{0.4}				& 48.7\exstd{0.4} \\
Bootstrap										& 72.8\exstd{0.4} 				& 66.5\exstd{0.4} 			& 57.4\exstd{0.7} 			& {\bf 13.4\exstd{0.9}} 		& 71.1\exstd{0.2} 		& 67.4\exstd{0.2} 			& 63.4\exstd{0.4} 				& 48.8\exstd{0.5} \\
Bootstrap + Mixup 								& 71.7\exstd{0.2} 				& 64.6\exstd{0.5} 			& 53.2\exstd{0.4} 			& 7.6\exstd{0.8} 				& 69.9\exstd{0.6} 		& 66.7\exstd{0.1} 			& 62.0\exstd{0.3} 				& 42.2\exstd{0.5} \\
Label smoothing 								& 73.0\exstd{0.2} 				& 67.7\exstd{0.4} 			& 60.6\exstd{0.2} 			& 9.1\exstd{0.4} 				& 71.2\exstd{0.5} 		& 68.2\exstd{0.5} 			& 64.2\exstd{0.3} 				& 50.2\exstd{0.4} \\
Label smoothing + Mixup 						& 72.5\exstd{0.5} 				& 67.9\exstd{0.4} 			& 60.9\exstd{0.6} 			& 6.8\exstd{0.1} 				& 71.1\exstd{0.3} 		& 68.3\exstd{0.3} 			& 63.8\exstd{0.1} 				& 47.3\exstd{0.1} \\
\midrule
\multicolumn{9}{c}{\textbf{\textsc{Ours}}}  \\ \midrule
\textbf{Ours:} NCR 				        & {\bf 73.4\exstd{0.5}}		& 67.8\exstd{0.3}				& 60.6\exstd{0.5}					& 11.5\exstd{0.5}					& {\bf 72.1\exstd{0.4}}		& {\bf 69.0\exstd{0.5}}			& {\bf 64.6\exstd{0.2}}			& {\bf 51.2\exstd{0.4}}	 			\\
\textbf{Ours:} NCR + Mixup 				& 73.1\exstd{0.3} 			& {\bf 68.3\exstd{0.1}} 		& {\bf 61.4\exstd{0.3}} 			& 7.1\exstd{0.4} 					& 71.7\exstd{0.4} 		& 68.6\exstd{0.2}					& 64.5\exstd{0.4} 				& 48.9\exstd{0.6} \\
\midrule
\multicolumn{9}{c}{\textbf{\textsc{Other works}}}  \\ \midrule
D-Mix~\citep{li2020dividemix}			& -- & -- & -- & -- & 55.8			& 50.3				& 50.9			& 35.4		\\
ELR~\citep{liu2020early}				& -- & -- & -- & -- & 57.4			& 58.1				& 50.6			& 41.7		\\
MOIT~\citep{ortego2020multi} 			& -- & -- & -- & -- & 64.7			& 63.1				& 60.8			& 45.9 		\\
\midrule
\multicolumn{9}{c}{\textbf{\textsc{Oracle: Clean subset}}}  \\ \midrule
Standard					& 72.7\exstd{0.4} 	& 70.4\exstd{0.1} 	& 67.2\exstd{0.3}			& 50.6\exstd{0.3}				& 70.9\exstd{0.5} 	& 68.6\exstd{0.5}		& 64.8\exstd{0.7}			& 47.5\exstd{0.5}  			\\
Mixup 						& 72.6\exstd{0.2}	& 70.3\exstd{0.3}	& 67.3\exstd{0.3}			& 51.8\exstd{0.4}				& 70.5\exstd{0.2}	& 68.6\exstd{0.3}		& 65.6\exstd{0.3} 			& 48.7\exstd{0.3} 			\\
\bottomrule
\end{tabular}
\end{center}

  \label{tab:baseline}
\vspace{-0.5\baselineskip}
\end{table*}

We now compare NCR against the baselines defined in Section~\ref{sec:relation}.
The results are reported on the official validation set of mini-ImageNet-\{Red, Blue\} datasets.
We run each experiment five times and report the mean accuracy at the completion of training.
We do not report the peak validation accuracy attained during training as the conclusions may be less likely to generalize to an unseen test set.

Table~\ref{tab:baseline} shows the final accuracy for each method on the mini-ImageNet-\{Red, Blue\} datasets across different noise splits.
When compared with the \emph{standard} baseline (eq.~\eqref{eq:loss}), our method significantly improves the performance, up to $4.9\%$ across all noise ratios.
Furthermore, we show that our method is compatible with some of the existing baselines.
Combining mixup with our method leads to further improvements in some scenarios.
We observe that NCR improves the accuracy of the method even at 0\% noise.
This suggests that it has a general regularization effect.
However, the improvement in accuracy is much more pronounced in the training sets which contain label noise.

\begin{figure*}
\centering
\scriptsize
\begin{tabular}{cc@{}c@{}c@{}c@{}}
& Blue-40\% & Blue-80\% & Red-40\% & Red-80\% \\
{\scriptsize Standard} &
\raisebox{-.5\height}{\includegraphics[width=25mm]{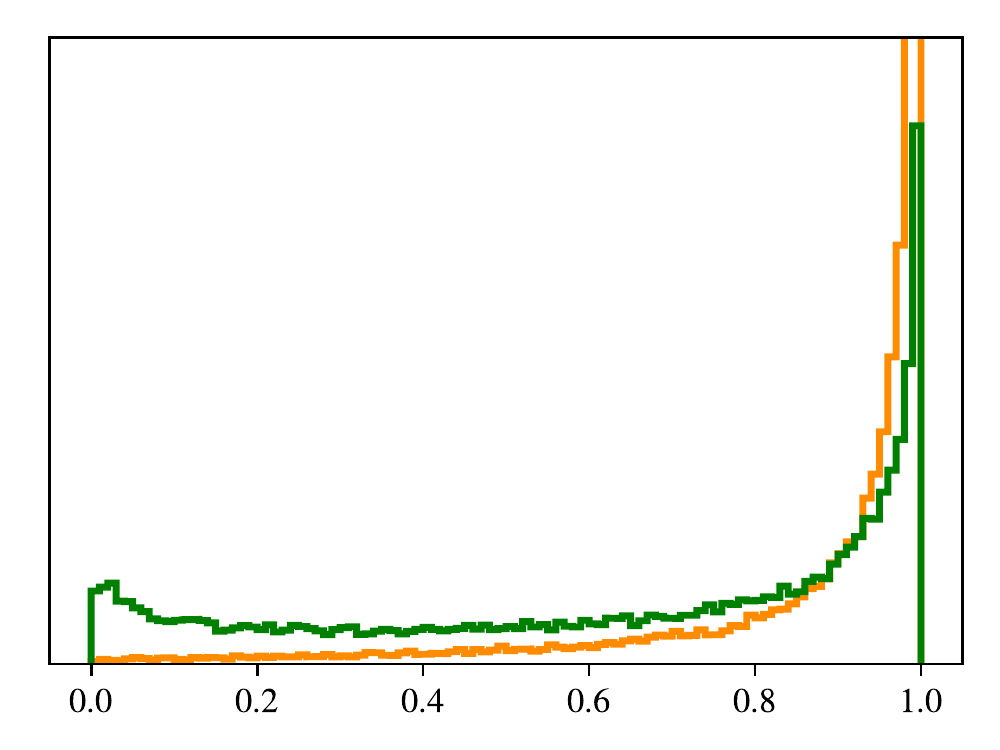}} &
\raisebox{-.5\height}{\includegraphics[width=25mm]{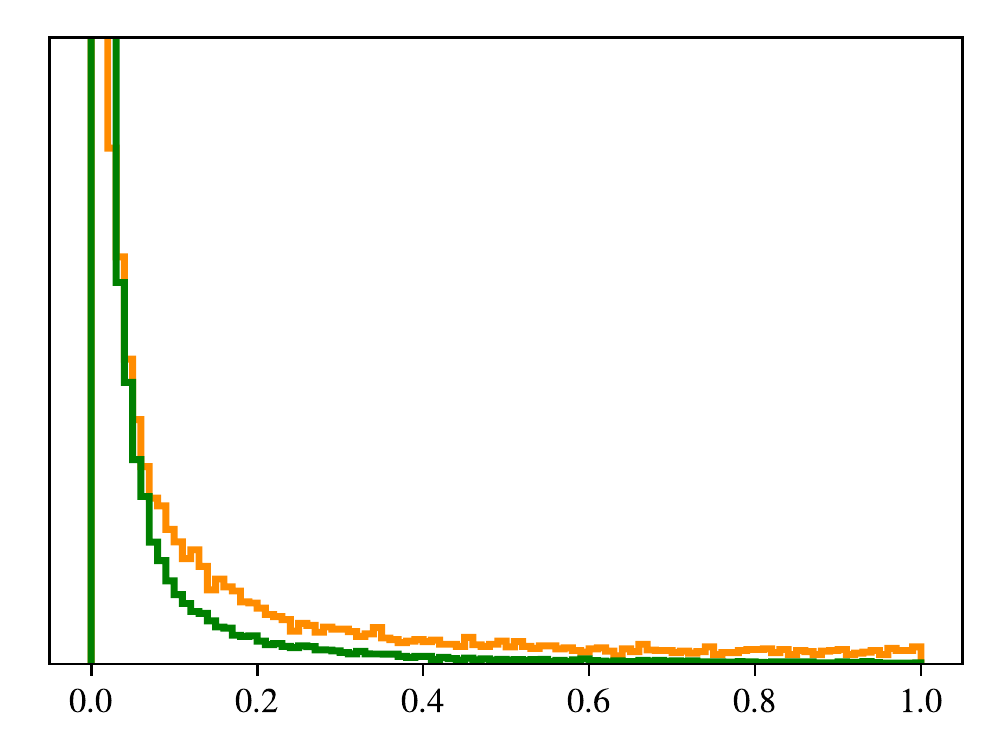}} &
\raisebox{-.5\height}{\includegraphics[width=25mm]{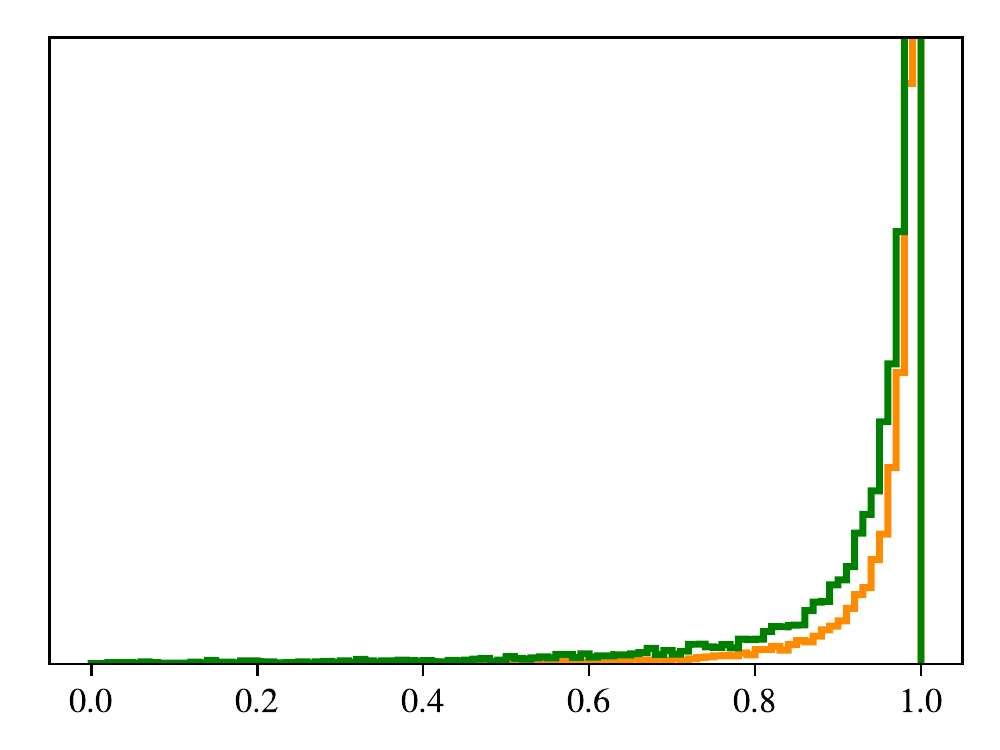}} &
\raisebox{-.5\height}{\includegraphics[width=25mm]{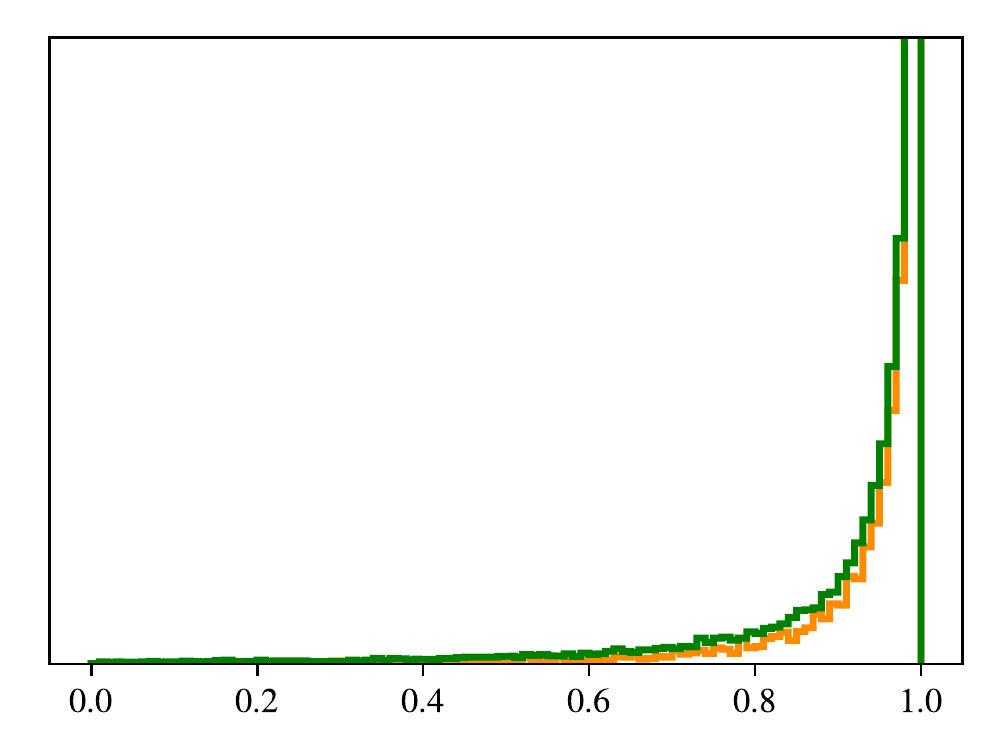}} \\
{\scriptsize NCR} &
\raisebox{-.5\height}{\includegraphics[width=25mm]{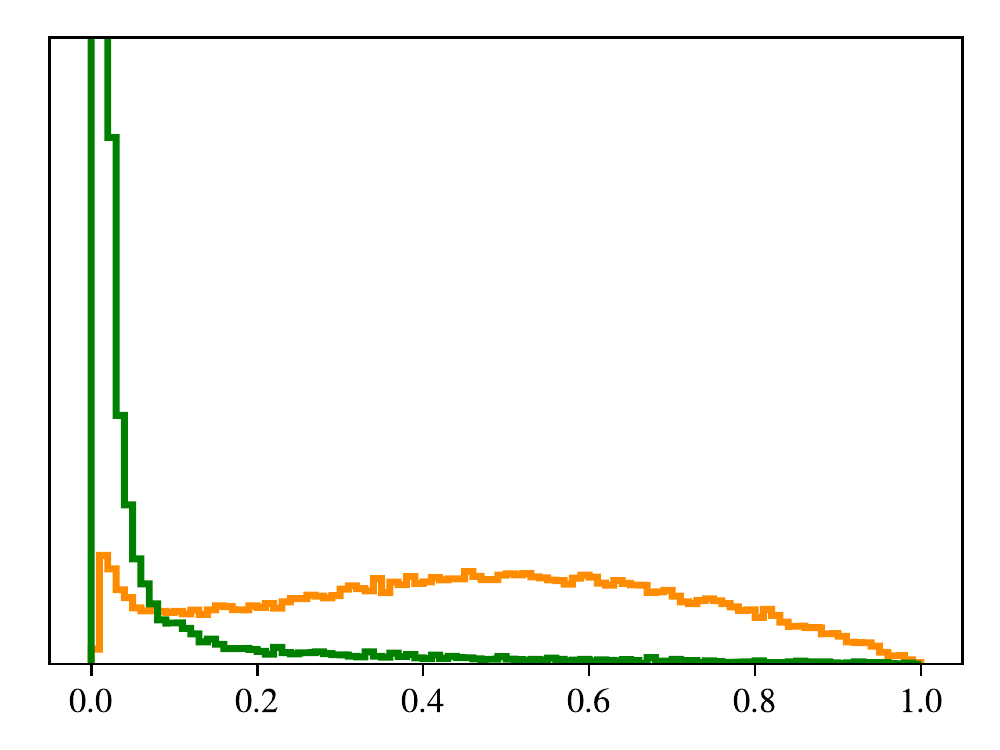}} &
\raisebox{-.5\height}{\includegraphics[width=25mm]{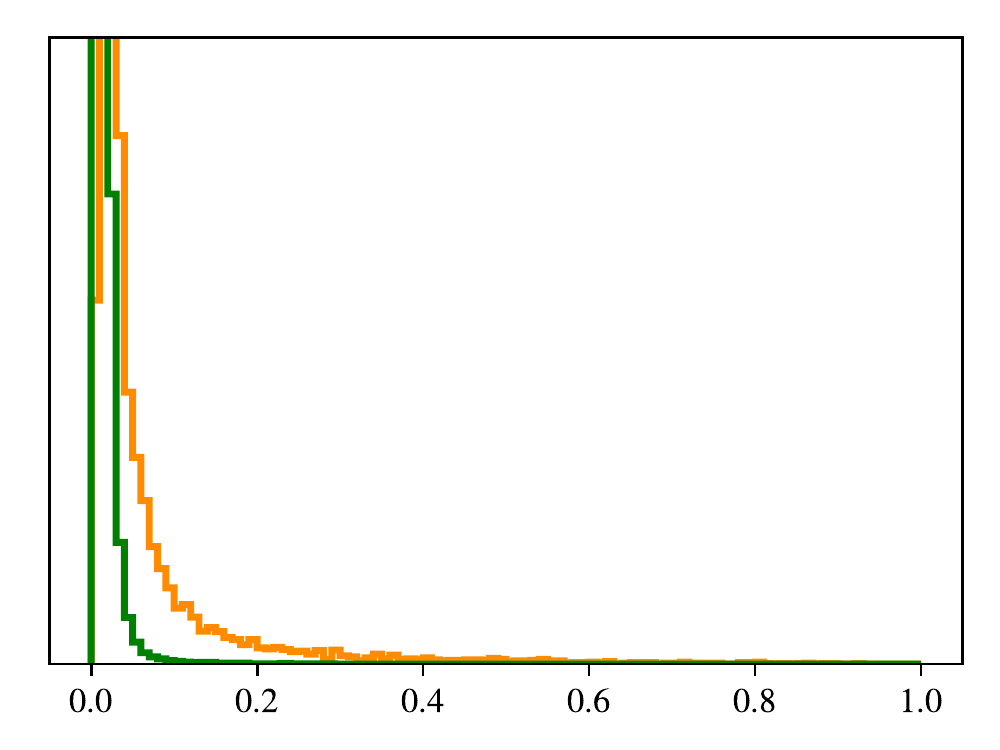}} &
\raisebox{-.5\height}{\includegraphics[width=25mm]{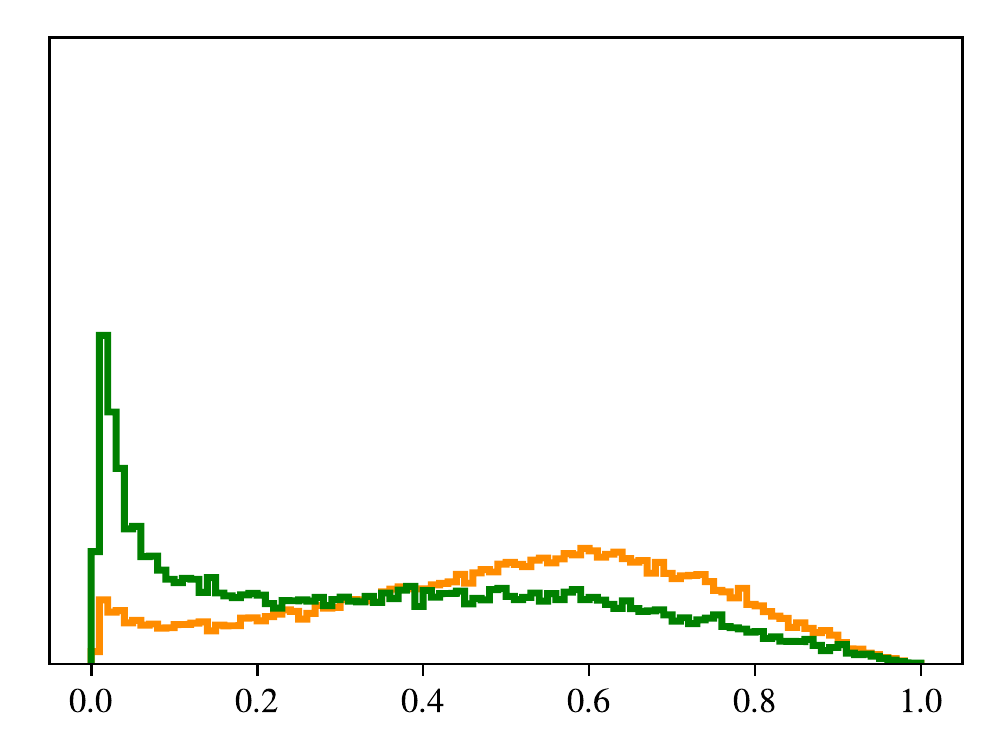}} &
\raisebox{-.5\height}{\includegraphics[width=25mm]{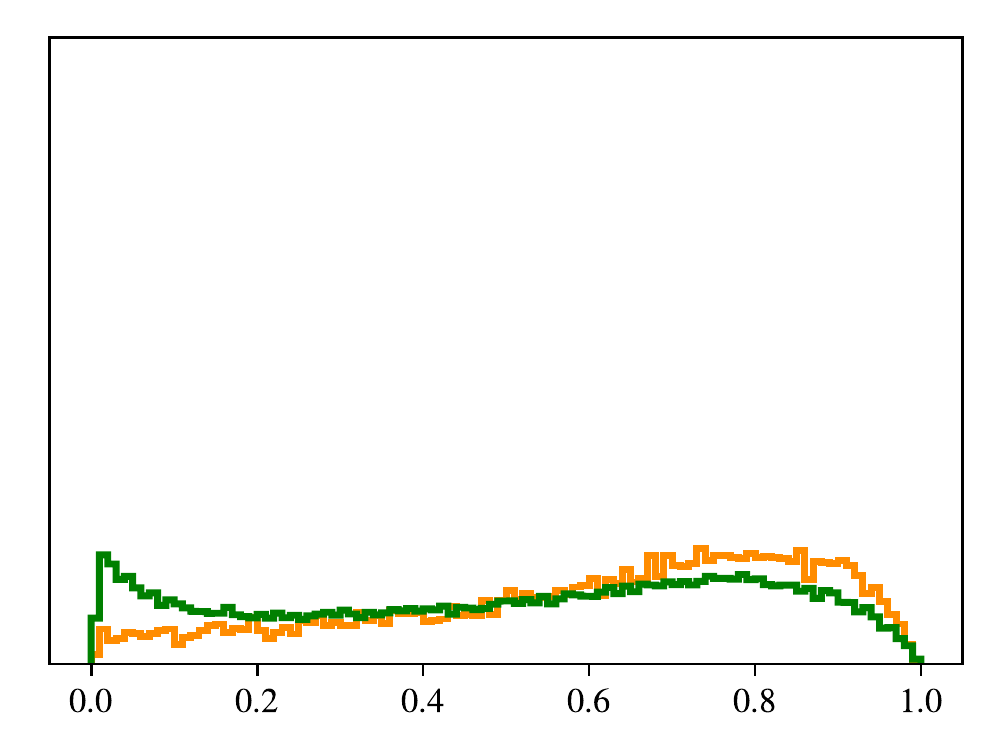}} \\
& \multicolumn{4}{c}{\includegraphics[scale=0.5]{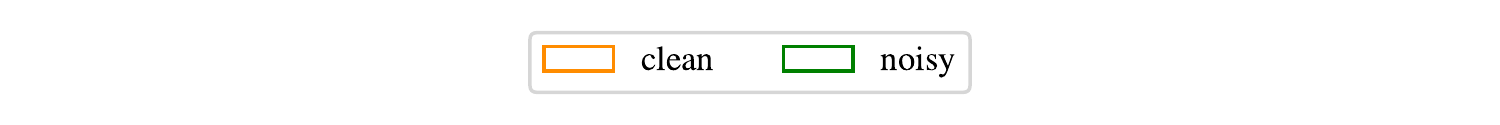}}
\end{tabular}
\vspace{-1\baselineskip}
\caption{\textbf{Predicted confidence of training examples.} Whereas the baseline assigns similar confidence scores to clean and mislabelled examples (usually all high), NCR more often assigns lower confidence to mislabelled examples and higher confidence to correct examples.
}
\label{fig:confidence} 
\end{figure*}

Figure~\ref{fig:confidence} presents further evidence that NCR inhibits memorization of noisy labels.
After the training is complete, we perform one last forward-pass and obtain the confidence $p$ that the model assigns to the annotated label for each training example.
The top row shows that the baseline model overfits to the noisy labels, resulting in $p=1$ for both clean and noisy images.
On the other hand, NCR avoids overfitting and assigns low confidence $p=0$ to most of the noisy training labels.
Some of the noisy examples are still classified as their assigned label on mini-Imagenet-Red. 
This is likely due to the dataset containing realistic and correlated noise:
the mislabelled class and the true class often have visual patterns in common.

For the synthetic noise (mini-Imagenet-Blue), NCR separates the clean and noisy examples up to $40\%$ noise ratio.
However, the model underfits with $80\%$ noise, resulting in $p$ being close to $0$ for both clean and noisy examples.
This results in a small improvement for NCR, see mini-ImageNet-Blue with $80\%$ noise in Table~\ref{tab:baseline}.

\begin{figure*}
\vspace{-0.5\baselineskip}
\centering
\scriptsize
\begin{tabular}{cc@{}c@{}c@{}c@{}}
& Blue-40\% & Blue-80\% & Red-40\% & Red-80\% \\
Standard &
\raisebox{-.5\height}{\includegraphics[width=25mm]{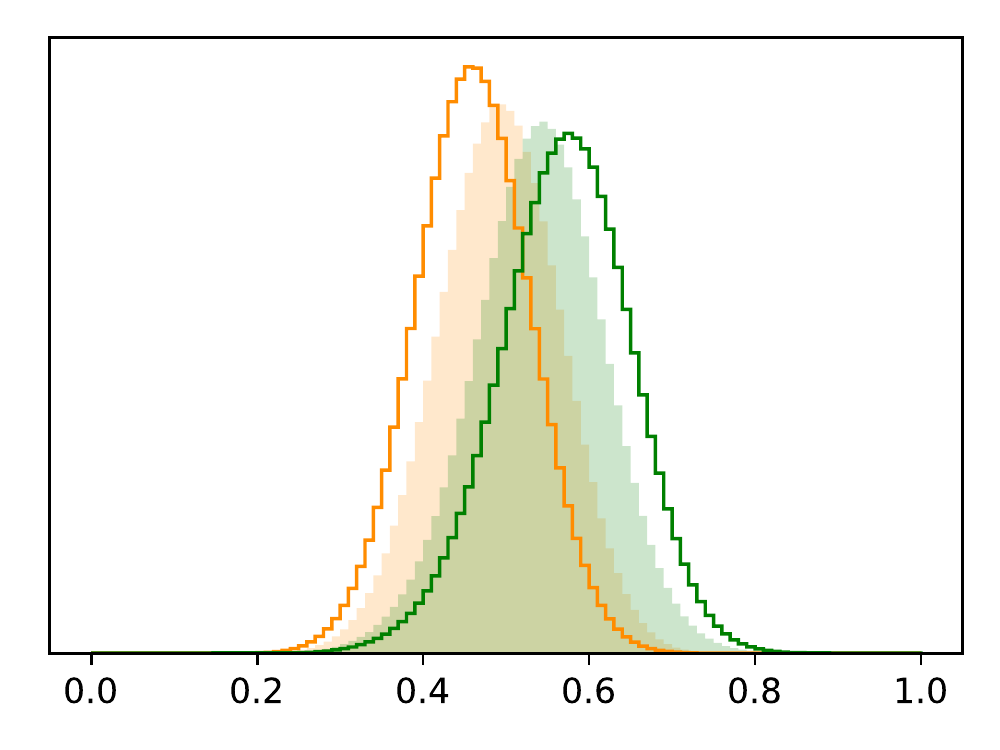}} &
\raisebox{-.5\height}{\includegraphics[width=25mm]{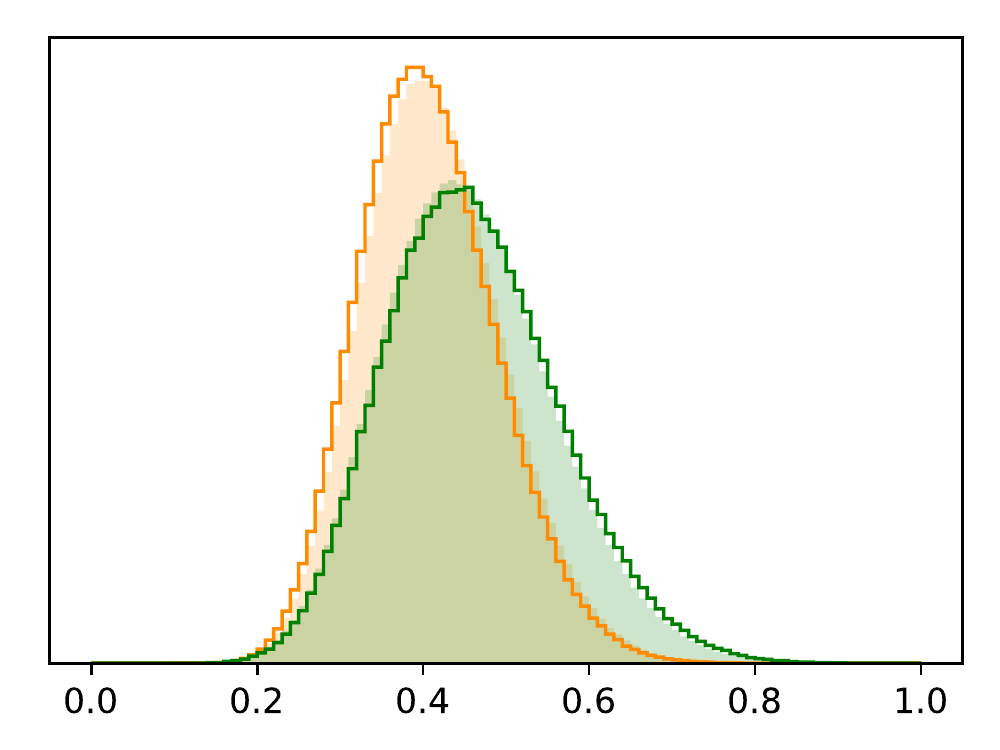}} &
\raisebox{-.5\height}{\includegraphics[width=25mm]{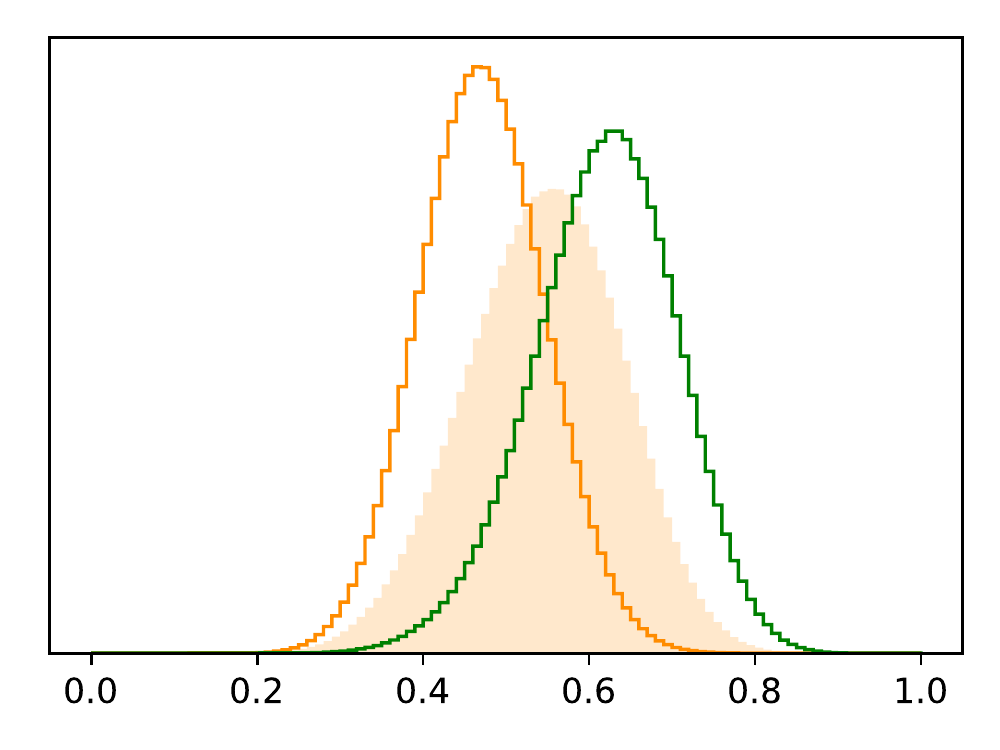}} &
\raisebox{-.5\height}{\includegraphics[width=25mm]{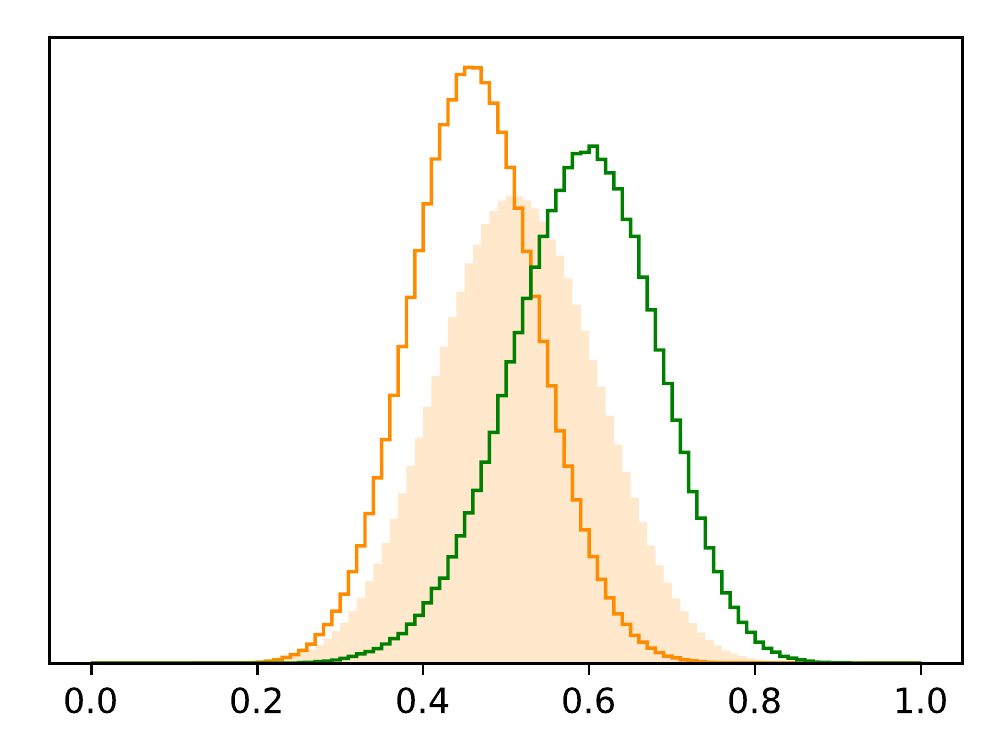}} \\
NCR &
\raisebox{-.5\height}{\includegraphics[width=25mm]{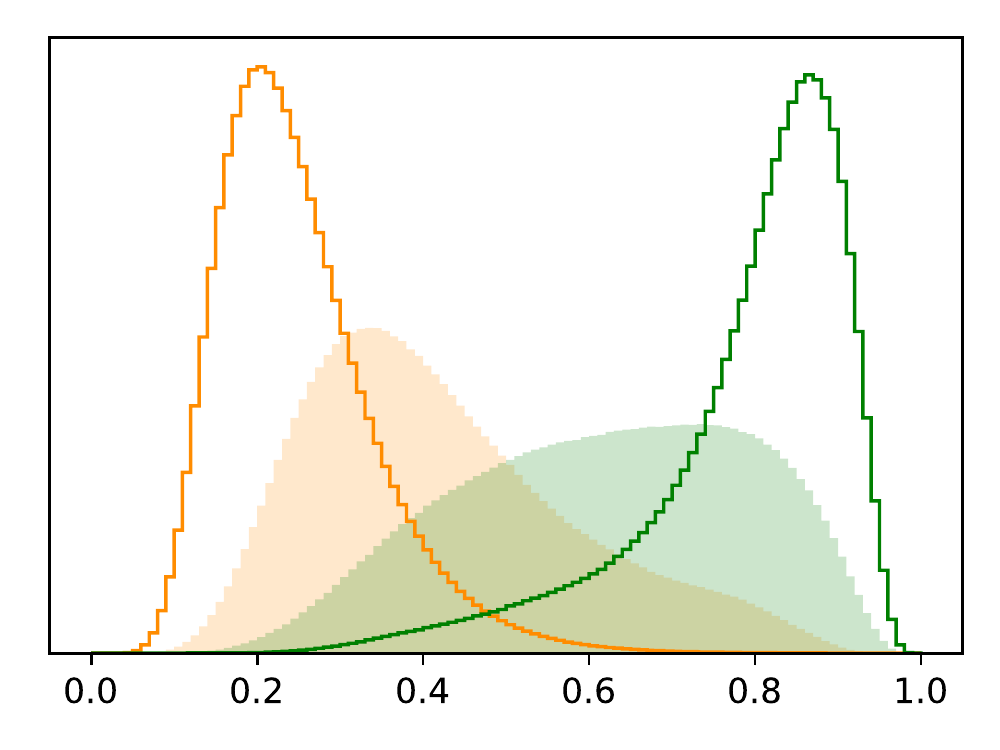}} &
\raisebox{-.5\height}{\includegraphics[width=25mm]{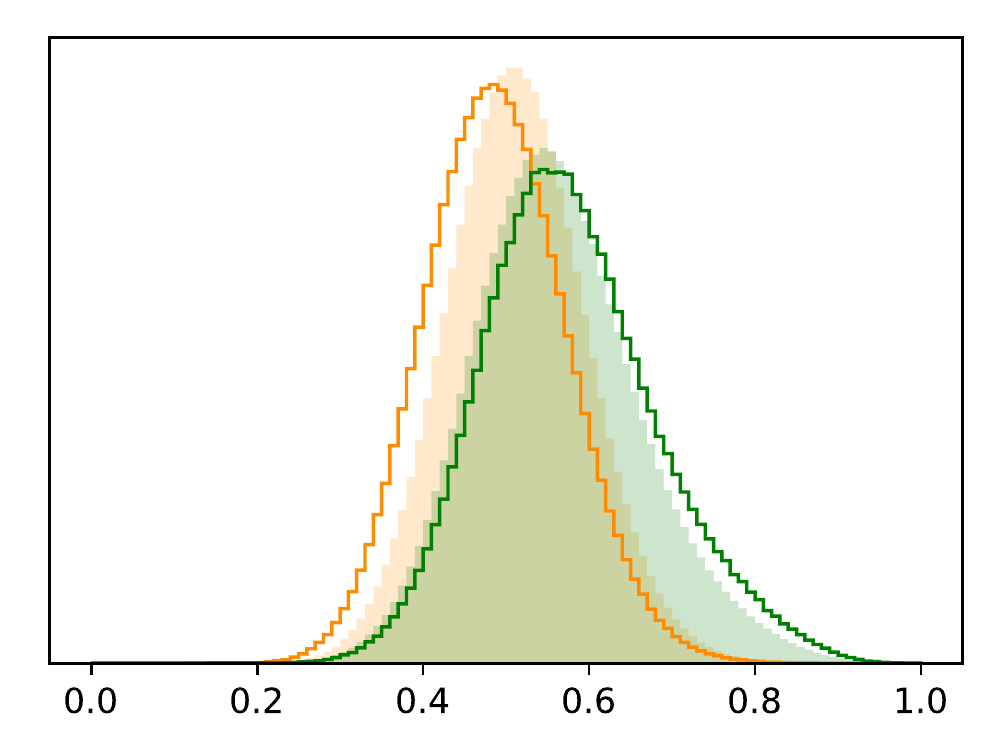}} &
\raisebox{-.5\height}{\includegraphics[width=25mm]{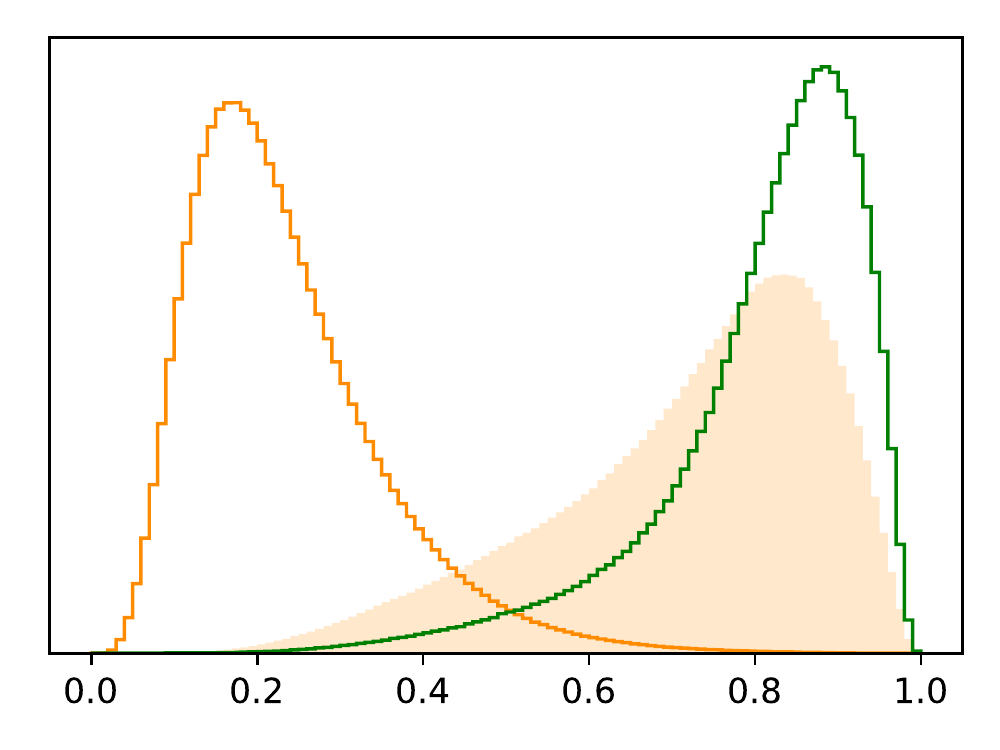}} &
\raisebox{-.5\height}{\includegraphics[width=25mm]{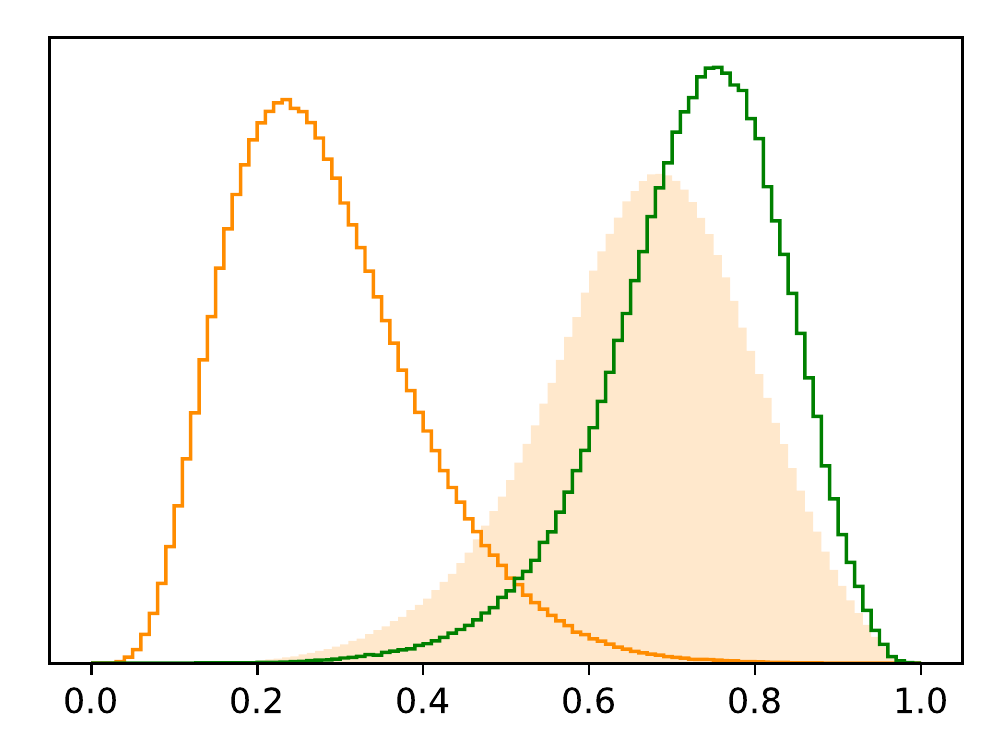}} \\
& \multicolumn{4}{c}{\includegraphics[scale=0.5]{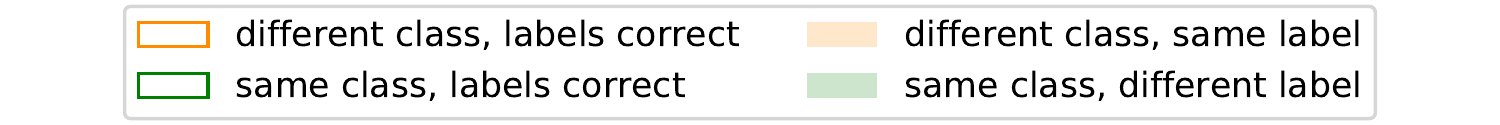}}
\end{tabular}
\caption{
\textbf{Similarity distributions.}
We compare the distribution of cosine similarities for training examples in mini-ImageNet that are correctly and incorrectly labelled as the same class or different classes.
For mini-ImageNet-Blue, the features learned using NCR achieve significantly better class separation with 40\% noise (or less, not pictured).
For the more realistic mini-ImageNet-Red, NCR still achieves better separation of the clean examples but fails to separate examples that are incorrectly labelled as the same class.
\label{fig:similarity}
}
\vspace{-0.5\baselineskip}
\end{figure*}

Table~\ref{tab:baseline} also includes an \emph{oracle} that is obtained by excluding the mislabelled examples from the training set, reducing its size by $20\%$, $40\%$ or $80\%$ accordingly.
Under \emph{realistic} noise (mini-ImageNet-Red), the results show that NCR outperforms the oracle across all noise ratios except for $40\%$ noise.
While this may be surprising, it can be explained by the observation that the noisy examples for each class are often visually similar to the clean examples, and thus still contain some useful information.
However, the performance of NCR is significantly less than the oracle under synthetic noise (mini-ImageNet-Blue), where the noisy examples are often entirely dissimilar to the clean examples.

\subsection{Effect on feature embeddings}

Using the datasets with known noise, we can compare the feature similarity of training examples that are correctly or incorrectly labelled as belonging to the same or different classes.
In the ideal case, the distributions of within-class and between-class similarities for clean examples would have zero overlap, and would be identical to the true within-class and between-class similarities for mislabelled examples.
Figure~\ref{fig:similarity} presents the similarity distributions for mini-ImageNet-Blue and -Red.
(Note that, for mini-ImageNet-Red, the set of clean examples is known but the true class of the mislabelled examples is not known, so we cannot obtain the true within-class similarity for mislabelled examples.)
We compare the distributions for a baseline model and a model trained with NCR.
While the distributions for the baseline model overlap, they are not identical, showing that the feature similarities contain some signal for NCR to take advantage of.
Training with NCR is shown to achieve greater separation of the classes in feature space, which must be due to the backpropagation of gradients through the feature similarities in~\eqref{eq:lcloss}.

\begin{table}
  \caption{
  \textbf{State-of-the-art comparison with realistic noise.}
  We compare NCR and our baselines to other methods on mini-WebVision, WebVision and Clothing1M. All results use ResNet-50, except for those marked by $^\dagger$ which use Inception-ResNetV2.
  }
  \small{
    \scalebox{0.93}{  
        \centering
        \begin{tabular}{@{\ssp}l@{\ssp}c@{\ssp}c@{\ssp}c@{\ssp}}
        \toprule
                                                                & mini-WebVision    & WebVision             & Clothing1M \\
        \midrule
        Standard                                                & 75.8              & 74.9                  & 71.7  \\ 
        Mixup                                                   & 77.2              & 75.5                  & 72.2   \\
        \textbf{Ours:} NCR                                      & 77.1              & 75.7                  & 74.4 \\
        \textbf{Ours:} NCR+Mixup                                & 79.4              & 75.4                  & 74.5 \\
        \textbf{Ours:} NCR+Mixup+DA                             & {\bf 80.5}        & {\bf 76.8}            & 74.6  \\
        \midrule
        MLNT; 3 iter.~\cite{li2019learning}                     & --                & --                       & 73.5 \\
        CleanNet~\cite{lee2018cleannet}                         & --                & --                       & {\bf 74.7}  \\
        $\mathcal{L}_{\textit{DMI}}$~\cite{wang2019derivative}  & --                & --                       & 72.5 \\
        LongReMix~\cite{cordeiro2021longremix}                  & --                & --                       & 73.0 \\ 
        ELR~\cite{liu2020early}                                 & 76.3$^{\dagger}$  & --                       & -- \\
        ELR+~\cite{liu2020early}                                & 77.8$^{\dagger}$  & --                       & {\bf 74.8} \\
        DMix~\cite{li2020dividemix}                             & 76.3              & --                       & {\bf 74.8} \\
        GJS~\cite{englesson2021generalized}                     & 79.3              & --                       & --  \\
        MoPro~\cite{li2020mopro}                                & --                & 73.9                  & --  \\
        MILe~\cite{rajeswar2021multi}                           & --                & 75.2                  & --  \\
        Heteroscedastic~\cite{collier2021correlated}            & --                & 76.6$^{\dagger}$      & --  \\
        CurrNet~\cite{guo2018curriculumnet}                     & --                & 79.3$^{\dagger}$      & --  \\
    
        \bottomrule
        \end{tabular}
    }
}

  \label{tab:sota}
\end{table}

\begin{table}
  \caption{
  \textbf{State-of-the-art comparison with synthetic noise on CIFAR.}
  A-40\% refers to 40\% asymmetric noise. All of the other columns refer to symmetric noise.
  }
  \makebox[\linewidth][c]{
\scriptsize
\scalebox{0.90}{
    \begin{tabular}{@{\ssp} l@{\ssp}| c@{\msp}c@{\ssp}c@{\ssp}c@{\ssp}c@{\ssp}c@{\msp}| c@{\ssp}c@{\ssp}c@{\ssp}c@{\ssp}c@{\ssp}c@{\ssp}}
    \toprule
    & \multicolumn{6}{c|}{CIFAR-10} & \multicolumn{6}{c}{CIFAR-100} \\
                & 20\%  & 40\%  & 50\%  & 80\%  & 90\%  & A-40\%    & 20\%  & 40\%  & 50\%  & 80\%  & 90\%  & A-40\% \\ \midrule
    Standard    & 83.9  & 68.3  & 58.5  & 25.9  & 17.3  & 77.3      & 61.5  & 46.2  & 37.4  & 10.4  & 4.1  & 43.9 \\
    MOIT+~\cite{ortego2020multi} & 94.1  & 92.0  & -     & 75.8  & -     & \textbf{93.2}      & 75.9  & 67.4  & -     & 51.4  & -     & 74.0 \\
    D-Mix~\cite{li2020dividemix}   & 95.1  & 94.2  & 93.6  & 91.4  & 74.5  & 91.8      & \textbf{76.7}  & 74.6  & \textbf{73.1}  & 57.1  & 29.7  & 72.1 \\
    ELR+~\cite{liu2020early}        & 94.9  & 94.4  & 93.9  & 90.9  & 74.5  & 88.9      & 76.3  & 74.0  & 72.0  & 57.2  & \textbf{30.9}  & 75.8 \\
    Ours+~\cite{liu2020early}   & \textbf{95.2}  & \textbf{94.5}  & \textbf{94.3}  & \textbf{91.6}  & \textbf{75.1}  & 90.7      & 76.6  & \textbf{74.2}  & 72.5  & \textbf{58.0}  & 30.8  & \textbf{76.3} \\   

    \bottomrule
    \end{tabular}
}

}

  \label{tab:sota_cifar}
\end{table}

\subsection{State-of-the-art comparison}

In addition to Table~\ref{tab:baseline}, which compares to prior art on mini-ImageNet-Red,
we compare NCR against the state-of-the-art on mini-WebVision, WebVision and Clothing1M (which evaluate realistic noise) in Table~\ref{tab:sota}.
Finally, we also compare to prior work on CIFAR-10 and -100 (which represents synthetic noise) in Table~\ref{tab:sota_cifar}.

In Table~\ref{tab:sota}, we also present a Data Augmentation (DA) variant of NCR, where we generate a second view of each example in the minibatch by applying random color jittering, motivated by~\cite{hoffer2020augment}.
Table~\ref{tab:sota} shows that NCR achieves state-of-the-art accuracy on datasets with realistic noise such as mini-WebVision, improving by 1.2\% on the best previous result.
NCR outperforms other methods on WebVision when ResNet50 is used, and is competitive against methods using Inception-ResNetV2, a much stronger architecture.
Our method is also competitive on Clothing1M, where it is only $0.2\%$ less than the state-of-the-art.

Note that NCR does not involve additional steps, such as dividing the dataset into multiple splits, learning multiple models, applying semi-supervised learning, or a second-stage with additional training as done in the other works~\cite{li2020dividemix, liu2020early, ortego2020multi} we compare to.
While NCR is orthogonal to these strategies, we show that we achieve state of the art results with minimal additional processing (\eg data augmentation).
Furthermore, NCR achieves higher performance compared to GJS~\citep{englesson2021generalized} which applies a similar consistency regularization only on different augmentation of each example.
This confirms that the neighbor consistency brings further improvements on top of augmentation consistency.

To show that our method is compatible with the existing approaches, we combine NCR with ELR~\cite{liu2020early} for CIFAR-10 and -100 comparisons in Table~\ref{tab:sota_cifar}.
We reproduce all the comparisons (by running the public code), except for MOIT~\cite{ortego2020multi} whose results are taken from the paper.
Our method consistently improves over ELR, achieving state-of-the-art results in almost all noise ratios in CIFAR.
For the noise ratios in which NCR does not obtain the highest result, it is typically the second-highest.
Note that we fix the same hyperparameters across all noise ratios, unlike Divide-Mix~\cite{li2020dividemix}, which is a more realistic scenario in practice.
\section{Conclusion} 
\label{sec:conclusion}

This work introduced Neighborhood Consistency Regularization and demonstrated that it is an effective strategy for deep learning with label noise.
While our approach draws inspiration from multi-stage training procedures for semi-supervised learning that employ transductive label propagation, it consists of a comparatively simple training procedure, requiring only that an extra loss be added to the objective which is optimized in stochastic gradient descent.
The efficacy of NCR is emphasized by the fact that it achieved state-of-the-art results under both synthetic (CIFAR-10 and -100) and realistic (mini-WebVision) noise scenarios.

\head{Limitations and future work.}
A limitation of NCR is that our proposed loss assumes that it has access to an adequate feature representation of the training data.
We overcome this limitation in practice by first training the network for $e$ epochs before applying the NCR loss, but future work is to remove this additional training hyperparameter.

Promising directions for future research including coupling NCR with a technique to rejecting out-of-distribution examples during training, as employed by other approaches in the literature~\cite{li2020dividemix,ortego2020multi}.
We also note that NCR could be applied to related problems, such as semi-supervised learning, as a regularization term.

\head{Broader Impact.}
Our proposed method is suited to learning from noisy data, as could be obtained by automatic scraping of the internet (illustrated by our experiments on mini-ImageNet-Red and mini-WebVision).
Data collected in this fashion may contain bias \citep{birhane2021large, de2019does} and a method which is able to more effectively learn from such data may inadvertently cause these biases to be amplified.
Furthermore, when using training data automatically scraped from the web, it is possible that the data is being used for a purpose to which the original owner did not consent, potentially infringing on their privacy.

{\small
\bibliographystyle{ieee_fullname}
\bibliography{egbib}
}
\clearpage
\newpage

 \twocolumn[
 \centering
 \Large
 \textbf{Learning with Neighbor Consistency for Noisy Labels} \\
 \vspace{0.5em}Supplementary Material \\
 \vspace{1.0em}
 ] 

\begin{table}[hbt!]
  \caption{List of network hyperparameters used in our experiments.
  \label{tab:hyperparams}
}
  \scriptsize
\centering
\begin{tabular}{@{\ssp}l@{\ssp}|@{\ssp}c@{\ssp}c@{\ssp}c@{\ssp}c@{\ssp}}
\toprule
									            & CIFAR-\{10, 100\}		& mini-\{Red, Blue\}  & mini-Webvision	& Clothing1M			\\
\midrule 
Opt.							            	& \multicolumn{4}{c}{SGD}																\\
Momentum							            & \multicolumn{4}{c}{0.9}																\\
Batch							            	& 256					& 128					& 256			& 128					\\
LR 					           					& 0.1					& 0.1					& 0.1			& 0.002					\\
LR Sch.										& \multicolumn{4}{c}{cosine decay with linear warmup}									\\
Warmup											& \multicolumn{4}{c}{5}																	\\
Epochs											& 250					& 130					& 130			& 80					\\
Weight Dec.										& $5e-4$				& $5e-4$				& $1e-3$		& $1e-3$				\\
Arch.											& \multicolumn{2}{c}{ResNet-18} & \multicolumn{2}{c}{ResNet-50}				\\

\bottomrule
\end{tabular}

\end{table}

\begin{table}
  \caption{List of NCR hyperparameters used in our experiments.
  \label{tab:ncrparams}
}
  \scriptsize
\centering
\begin{tabular}{@{\ssp}l@{\ssp}|@{\ssp}l@{\ssp}l@{\ssp}l@{\ssp}l@{\ssp}|l@{\ssp}l|l@{\ssp}|l@{\ssp}|l@{\ssp}}
\toprule
			& \multicolumn{4}{c|}{mini-ImageNet}             & CIFAR-10 & CIFAR-100 & WebVision & Clothing1M \\
			& 0\% 		& 20\% 		& 40\% 		& 80\% 		 &  		& 			&			& \\ \midrule
$\alpha$ 	& $0.9$		& $0.9$		& $0.7$		& $0.9$		 & $0.1$	& $0.1$ 	& $0.5$		& $0.9$ \\
$k$			& $50$		& $10$		& $5$		& $100$		 & $10$		& $10$   	& $10$			& $1$   \\
$e$			& $50$		& $50$		& $50$		& $50$		 & $50$		& $200$  	& $0$			& $40$  \\
\bottomrule
\end{tabular}


\end{table}

\appendix

\setcounter{page}{1}


\section{Training details}
\label{sec:details}

\head{Implementation details. }
We use the ResNet-18 and -50 architectures \citep{he2016deep} in our experiments.
We follow the same protocol as ELR~\citep{liu2020early}, in CIFAR experiments. 
For Clothing1M, we follow the work of~\citet{li2020dividemix} and fine-tune a pre-trained ResNet-50 for 80 epochs, where each epoch contains 1000 mini-batches. 
Table~\ref{tab:hyperparams} lists the network hyperparameters used to train the network throughout our experiments.
We train the network with the typical dot-product linear classifier $ h\prm{W}()$ in all datasets except for mini-WebVision and WebVision.
For the mini-WebVision and WebVision experiments, we follow the work of~\citet{ortego2020multi} and use a \emph{cosine classifier} for $h\prm{W}()$. 
Cosine classifier is also a linear classifier, however, the features and the classifier weights are $\ell_2$-normalized unlike the dot-product classifier.

We employ random crop augmentation during training in all experiments and resize images to $224 \times 224$ pixels.
For experiments with CIFAR, we use $32 \times 32$ images and reduce the strides.
We trained each model on a single Nvidia V100 GPU, and will release all code upon acceptance.

\head{NCR hyperparameters. }
We sweep over the NCR hyperparameters $\alpha$, $k$ and $e$, and choose a set of hyperparameter based on the validation set accuracy on CIFAR-\{10, 100\} and Clothing1M.
This hyperparameter sweep is done for each noise ratio separately.
Since mini-ImageNet-\{Red, Blue\} does not contain a held-out validation set , we create a \emph{held-out} set from the mini-ImageNet-Red dataset which comprises the (clean) examples from the $0\%$ noise dataset that do not appear in the datasets with $20\%$, $40\%$ or $80\%$ noise.
The \emph{held-out} set allows us to choose hyperparameters without overfitting on the final evaluation set.
We use the same hyperparameters on mini-ImageNet-Blue as well.
Table~\ref{tab:ncrparams} shows the list of hyperparameters for each dataset.

\section{Dataset details}
\textbf{Mini-ImageNet-Red} contains 50 000 training examples and 5 000 validation examples. 
The noisy images are retrieved by text-to-image and image-to-image search.
They come from an open vocabulary outside of the set of classes in the training set.
Depending on the noise ratio, a subset of clean images are replaced by the noisy images to construct the training set.
\textbf{Mini-ImageNet-Blue} contains 60 000 training examples.
The validation set is the same as mini-ImageNet-Red.
The noise in mini-ImageNet-Blue is synthetic. 
The label of each example is independently and uniformly changed according to a probability.
The noisy examples come from a fixed vocabulary, \ie their true label belongs to another class in the training set.
\textbf{WebVision} contains 2.4M images and 1000 classes. 
Images are collected from the web, using the Google and Flickr search engines.
The data is \emph{imbalanced}, meaning each class contains a different number of training examples.
\textbf{Mini-Webvision} contains a subset of the original Webvision dataset~\cite{li2017webvision}.
It contains only the first 50 classes of the Google image subset.
This corresponds to 65 944 training images.
The validation set contains 2 500 images corresponding to the 50 training classes.
\textbf{Clothing1M}~\citep{xiao2015learning} is a large-scale dataset containing 1 million images and 14 categories. 
Images are collected from the web, and the noisy labels are assigned based on the surrounding text.
We do not use the clean training subset with human-verified labels.
We follow the existing protocol~
\citep{li2019learning} and fine-tune a ResNet-50 model which is pre-trained on ImageNet.

\section{Effect of batch size and number of labels}

\begin{table}[t]
\small
    \centering
    \caption{Effect of the batch size on our proposed NCR method, on the WebVision dataset containing 1000 classes.}
    \begin{tabular}{lcccc}
    \toprule
    Batch Size & 256 & 512 & 1024 & 2048 \\ 
    Accuracy   & 73.9 & 75.0 &  75.7 & 75.6 \\
    \bottomrule
    \end{tabular}
    \label{tab:webvision_rebuttal}
\end{table}

We use the WebVision dataset to to analyse the effect of the batch size for large label vocabularies.
WebVision is a large-scale dataset containing $2.4$M images and $1$K classes.
We report the accuracy for different batch sizes in Table~\ref{tab:webvision_rebuttal}.
The accuracy increases with batch size, plateauing after the batch size is more than the number of classes.
This shows that NCR requires the batch size to be approximately the number of classes, but that much bigger batch sizes are not required.
We use a batch size of $1024$ for WebVision. 
The other training hyperparameters remain the same as mini-Webvision on Table~\ref{tab:hyperparams}.

\end{document}